\documentclass[letterpaper]{article} 
\usepackage[preprint]{aaai2027} 
\usepackage[hyphens]{url} 
\usepackage{graphicx} 
\usepackage{natbib} 
\usepackage{caption} 
\usepackage{amsmath}
\usepackage{amssymb}
\usepackage{booktabs}
\usepackage{xcolor}

\newcommand{\sg}{\operatorname{sg}}
\newcommand{\method}{\textsc{UniCycleFlow}}
\definecolor{externalgray}{gray}{0.48}
\newcommand{\best}[1]{\textcolor{black}{\textbf{#1}}}
\newcommand{\second}[1]{\textcolor{black}{\underline{#1}}}
\newcommand{\ext}[1]{\textcolor{externalgray}{#1}}

\title{UniCycleFlow: Bidirectional Unpaired Image Translation with a Shared Rectified Flow}

\makeatletter
\@ifundefined{name}{\newcommand{\name}[1]{\author{#1}}}{}
\@ifundefined{address}{\newcommand{\address}[1]{\affiliations{#1}}}{}
\makeatother

\name{
    Xianhao Zhou$^1$, Jianghao Wu$^2$, Shaoting Zhang$^{1,3}$, Guotai Wang$^{1,3,*}$
}
\address{
$^1$ School of Mechanical and Electrical Engineering,\\
University of Electronic Science and Technology of China, Chengdu, China\\
$^2$ Faculty of Information Technology, Monash University, Melbourne, Australia\\
$^3$ Shanghai Artificial Intelligence Laboratory, Shanghai, China\\
$^*$ Corresponding author: Guotai Wang
(\mbox{\texttt{guotai.wang@uestc.edu.cn}})
}

\begin{document}
\maketitle

\begin{abstract}
Bidirectional unpaired image translation must preserve source-specific structure while learning coherent transformations in both directions without paired supervision. Existing methods typically employ two direction-specific generators or train separate one-way models. Even when linked by cycle consistency, such models constrain only the round-trip endpoint reconstruction, without requiring the two directions to obey a common local transformation rule. We propose UniCycleFlow, a rectified-flow framework that represents bidirectional translation as forward and reverse integration of a single time-conditioned velocity field. This formulation organizes both directions within the same continuous dynamics, rather than coupling otherwise separate endpoint mappings. A key challenge is that unpaired data provide no meaningful source--target coupling from which rectified-flow trajectories can be constructed. UniCycleFlow addresses this challenge by learning deterministic source-conditioned endpoints whose marginal distributions are adversarially matched to the opposite domains. The resulting paths are regularized by stop-gradient self-flow matching for intermediate velocity supervision, discrete cycle closure for forward--reverse consistency, and representation path-velocity regularization for controlling localized feature changes along the trajectory. Across ten translation directions, UniCycleFlow achieves the lowest FID on 7 of 10 tasks using a single Euler evaluation and obtains the best average FID of 55.1.

\end{abstract}

\section{Introduction}

Unpaired image-to-image translation aims to transform images across visual domains without aligned source--target pairs. In the absence of instance-level correspondence, matching the target-domain distribution alone is insufficient to determine how each source image should be translated, making source-structure preservation a central challenge. Existing methods address this ambiguity through cycle reconstruction, representation correspondence, spatial consistency, or pretrained generative priors. In bidirectional settings, the two directions are typically implemented by separate mappings or selected through conditional generation. Their relationship is therefore established mainly through endpoint-level objectives or discrete conditioning, rather than represented directly by a common transformation process.

Cycle-consistent methods jointly train two direction-specific generators and relate them through round-trip reconstruction \cite{zhu2017cyclegan,yi2017dualgan,kim2017discogan}. Although the cycle objective effectively encourages source preservation and consistency between the translated endpoints, the transformations performed by the individual generators remain independently parameterized. One-way correspondence-based methods instead preserve source information through contrastive or spatial constraints \cite{park2020cut,han2021dclgan,zheng2021spatial}, but applying them in both directions generally requires separate models or training runs. Conditional and shared-generator architectures reduce parameter duplication using domain labels or direction codes \cite{choi2018stargan,shen2020one}, but parameter sharing alone does not impose common continuous dynamics on the two directions.

UniCycleFlow instead assigns the two domains to opposite ends of a temporal interval and realizes translation by integrating one time-conditioned velocity field forward or backward. Both directions are thus governed by a common time-indexed transformation model, rather than implemented by separate generators or selected through a discrete direction condition. This formulation determines how the two directions are organized, but applying rectified flow to unpaired translation still requires cross-domain endpoint pairs from which training trajectories and velocity targets can be constructed.

Standard rectified flow constructs interpolation paths from samples drawn from a prescribed endpoint coupling. In the unpaired setting, independently sampling real images from the two domains produces displacements that combine the desired domain shift with unrelated differences in content, pose, and spatial layout. The resulting velocity targets are therefore poorly aligned with source-conditioned image translation. UniCycleFlow instead learns a deterministic endpoint for each source image and adversarially matches the distribution of the generated endpoints to the opposite domain. These source-conditioned endpoint pairs define translation-specific trajectories without relying on arbitrary pairings between unrelated real images.

The learned endpoints specify the overall cross-domain displacement, but endpoint supervision alone does not fully constrain the induced transport. Adversarial boundary matching does not directly supervise the intermediate states encountered during integration, while the discrete forward and reverse endpoint updates are not guaranteed to compose consistently. Moreover, a straight trajectory in image space may still exhibit excessive localized feature changes in a perceptual representation. UniCycleFlow therefore complements endpoint learning with trajectory-level regularization. Stop-gradient self-flow matching transfers the boundary displacement to intermediate field predictions, discrete cycle closure constrains the forward--reverse composition, and representation path-velocity regularization controls localized feature changes between neighboring trajectory states using a frozen dense encoder.

By combining shared temporal dynamics, source-conditioned trajectory construction, and path-level supervision, UniCycleFlow provides a unified rectified-flow framework for bidirectional unpaired translation. Across ten translation directions, UniCycleFlow achieves the lowest FID on 7 of 10 tasks under one-step inference and obtains the best ten-direction average FID and KID$\times100$ of 55.1 and 2.107, respectively. 

\section{Related Work}

\subsection{Unpaired and Bidirectional Image Translation}

Adversarial methods for unpaired image translation can be broadly grouped into cycle-consistent bidirectional mappings, latent-representation models, and conditional or shared-generator architectures. CycleGAN, DualGAN, and DiscoGAN jointly learn two direction-specific generators and relate them through round-trip reconstruction \cite{zhu2017cyclegan,yi2017dualgan,kim2017discogan}. UNIT instead assumes a shared latent space across domains \cite{liu2017unit}, while MUNIT decomposes images into domain-invariant content and domain-specific style to support multimodal translation \cite{huang2018munit}. Conditional approaches such as StarGAN and StarGAN v2 handle multiple domains within a unified architecture \cite{choi2018stargan,choi2020starganv2}, whereas One-to-one CycleGAN investigates a shared self-inverse mapping for two-domain translation \cite{shen2020one}.

Recent work has extended these paradigms by improving generator design, content--style modeling, and robustness to asymmetric domain correspondences. UVCGAN and UVCGAN v2 incorporate stronger hybrid architectures and training strategies while retaining cycle-based endpoint supervision \cite{torbunov2023uvcgan,torbunov2023uvcganv2}. SHUNIT harmonizes source and target styles to better preserve fine-grained source characteristics \cite{song2023shunit}, while StegoGAN addresses non-bijective translation, where conventional cycle-consistent models may hide unmatched information or introduce spurious target-domain content \cite{wu2024stegogan}. Despite their different designs, these methods relate the two directions mainly through endpoint reconstruction, latent structure, or discrete conditioning, leaving their local transformations only indirectly coupled. UniCycleFlow instead models them as opposite-time traversals of a single velocity field, directly coupling forward and reverse changes and providing a shared trajectory for path-level supervision.

\subsection{Content and Path Regularization}

Content-preservation objectives in unpaired translation mainly operate through endpoint correspondence or mapping-geometry regularization. Correspondence-based methods preserve source information by comparing source and output representations. CUT, DCLGAN, and LSeSim impose patch-level contrastive or spatial correspondence constraints \cite{park2020cut,han2021dclgan,zheng2021spatial}, while EnCo aligns latent encoder and decoder patches without an additional Siamese network \cite{cai2024enco}.

A complementary line of work regularizes the geometry of the learned
mapping. DistanceGAN preserves pairwise distances across domains
\cite{benaim2017one}, while GcGAN imposes consistency under
geometric transformations \cite{fu2019geometry}. Density-changing
regularization accounts for changes in probability density across
domains \cite{xie2022density}, and shortest-path regularization
(SANTA) favors mappings with shorter transport paths
\cite{xie2023shortest}. Correspondence objectives primarily compare the source and translated endpoints, whereas geometric objectives characterize the distributional or aggregate path properties of the mapping. Our representation path-velocity regularization (RPV) instead evaluates finite-difference feature changes between neighboring states of the model-induced continuous trajectory. By comparing corresponding locations in a frozen dense representation, RPV provides spatially resolved supervision for localized path variations that can be obscured by endpoint comparisons or globally pooled features.

\subsection{Diffusion, Bridge, and Flow-Based Translation}

Recent transport-based translation methods include diffusion editing and inversion, domain-specific diffusion transport, and Schr\"odinger-bridge models. SDEdit translates an input by adding noise and subsequently denoising it with a pretrained diffusion model \cite{meng2022sdedit}, while EGSDE introduces energy guidance for unpaired translation \cite{zhao2022egsde}. DDIB connects probability-flow ODEs learned separately for the two domains \cite{su2023ddib}, producing bidirectional translation through domain-specific diffusion processes rather than a directly learned shared cross-domain field. Schr\"odinger-bridge approaches instead learn transport between the source and target distributions: UNSB uses adversarial bridge learning for unpaired translation \cite{kim2024unsb}, and IBCD distills an implicit bridge into a single-step bidirectional model \cite{lee2025ibcd}. These approaches typically use iterative diffusion, multi-step stochastic transport, or bridge distillation.

Flow matching provides a continuous-time alternative by learning a vector field without simulating its ODE trajectory during optimization \cite{lipman2023flow}. Rectified flow further favors straight transport paths, allowing the learned field to be evaluated with a small number of numerical integration steps \cite{liu2023rectified}. Standard formulations, however, construct interpolation paths and velocity targets from a prescribed endpoint coupling. In unpaired translation, independently sampling real endpoints from the two domains mixes the desired domain shift with arbitrary inter-instance variation and therefore provides poorly aligned velocity supervision. UniCycleFlow learns deterministic source-conditioned endpoints through adversarial boundary matching, constructs translation-specific trajectories from the resulting pairs, and trains a single time-conditioned field shared by both translation directions.

\section{Method}

\subsection{Problem Formulation and Standard Rectified Flow}

Let $x_A\sim p_A$ and $x_B\sim p_B$ denote unpaired samples from domains $A$ and $B$, respectively. We associate domain $A$ with time $t=0$ and domain $B$ with time $t=1$, and parameterize a time-conditioned velocity field $v_\theta(x,t)$ through the ordinary differential equation
\begin{equation}
  \frac{d x_t}{dt}=v_\theta(x_t,t), \qquad t\in[0,1].
  \label{eq:method-ode}
\end{equation}

Rectified flow assumes an endpoint coupling $(x_0,x_1)\sim\pi$ and constructs a straight interpolant $x_t=(1-t)x_0+t x_1$. Since the corresponding path has constant velocity $x_1-x_0$, the standard training objective is
\begin{equation}
  \mathcal L_{\mathrm{RF}}
  =
  \mathbb E_{(x_0,x_1)\sim\pi,\,t}
  \left\|
  v_\theta(x_t,t)-(x_1-x_0)
  \right\|_1,
  \qquad
  t\sim\mathcal U(0,1).
  \label{eq:standard-rf}
\end{equation}

A direct application to unpaired translation independently samples $x_A\sim p_A$ and $x_B\sim p_B$ and uses $(x_0,x_1)=(x_A,x_B)$. Although this pairing defines a valid coupling of the two marginal distributions, its displacement $x_B-x_A$ combines the desired domain shift with arbitrary differences in content, pose, and spatial layout between unrelated images. The resulting velocity targets are therefore poorly aligned with source-conditioned image translation. We retain this independent-pair formulation as a controlled baseline in the ablation study, and replace its randomly paired endpoints with learned source-conditioned endpoints in UniCycleFlow.

\subsection{Shared Source-Conditioned Bidirectional Transport}

UniCycleFlow uses a single velocity field for both translation directions. The field always predicts velocity in the canonical $A\rightarrow B$ orientation; it does not choose between a forward and a reverse velocity according to the input. Translation direction is instead determined by the sign of the time increment. For $K$ Euler steps with $h=1/K$ and $t_k=kh$, the two directions are evaluated as
\begin{align}
  A\rightarrow B:\quad
  x_{k+1}
  &=x_k+h\,v_\theta(x_k,t_k),
  && k=0,\ldots,K-1,
  \nonumber\\
  B\rightarrow A:\quad
  x_{k-1}
  &=x_k-h\,v_\theta(x_k,t_k),
  && k=K,\ldots,1.
  \label{eq:bidirectional-integration}
\end{align}
Thus, at any intermediate state and time, the network output has the same canonical orientation. Forward translation follows the field with a positive time increment, whereas reverse translation traverses the same field with a negative increment.

A single Euler step gives the boundary endpoint maps used during training:
\begin{align}
  v_A &= v_\theta(x_A,0),
  &\hat{x}_B &= x_A+v_A,
  \nonumber\\
  v_B &= v_\theta(x_B,1),
  &\hat{x}_A &= x_B-v_B.
  \label{eq:endpoints}
\end{align}
Here, $v_A$ and $v_B$ denote boundary velocities. A hat denotes a generated endpoint in the subscripted domain: $\hat{x}_B$ is generated from $x_A$, whereas $\hat{x}_A$ is generated from $x_B$. These endpoint maps coincide with the model outputs under one-step inference.

Instead of pairing each source with an independently sampled real image, UniCycleFlow uses the deterministic pairs $(x_A,\hat{x}_B)$ and $(\hat{x}_A,x_B)$. Their generated endpoint marginals are constrained to match the opposite domains through adversarial boundary supervision, described later. Each pair then defines a source-conditioned straight trajectory and a translation-specific displacement for intermediate velocity training.

\begin{figure*}[t]
\centering
\includegraphics[width=\linewidth]{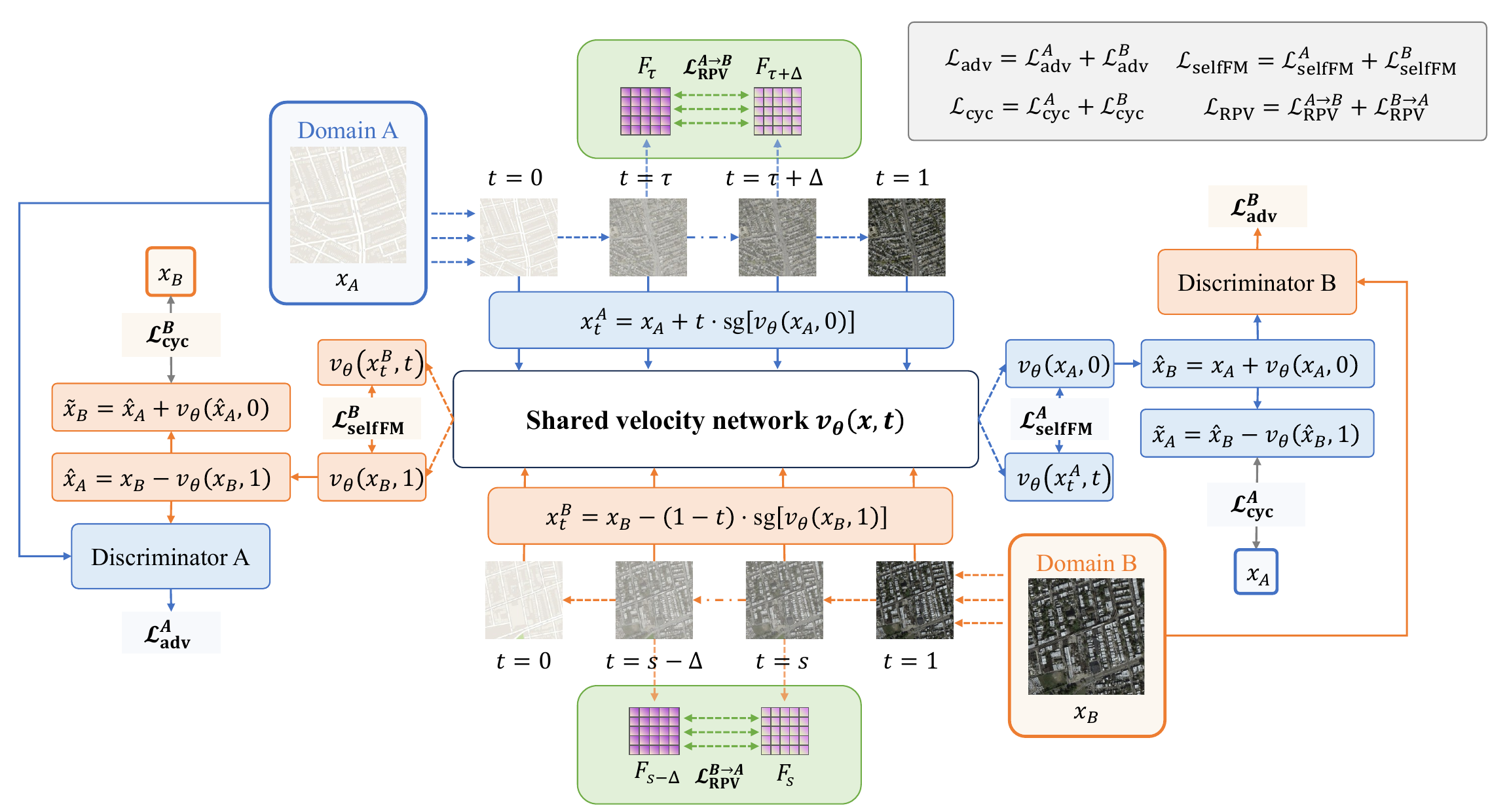}
\caption{Overview of UniCycleFlow. A single time-conditioned velocity field translates $A\rightarrow B$ by forward integration from $t=0$ and $B\rightarrow A$ by reverse integration from $t=1$. Generated endpoints are matched to the opposite-domain marginals. SelfFM supervises intermediate field predictions using detached boundary velocities, cycle closure regularizes the discrete forward--reverse composition, and RPV measures dense representation changes between neighboring states of each source-conditioned trajectory.}
\label{fig:overview}
\end{figure*}

\subsection{Trajectory Regularization}

The learned endpoint maps determine the overall cross-domain displacement and the generated target marginals, but do not fully constrain the induced transport. First, adversarial boundary supervision does not directly train the velocity field at intermediate states encountered during multi-step integration. Second, the discrete forward and reverse endpoint updates in Eq.~\eqref{eq:endpoints} are not guaranteed to compose to the identity. We address these issues using stop-gradient self-flow matching and discrete cycle closure, respectively.

\paragraph{Stop-gradient self-flow matching.}
For each direction, we sample $t\sim\mathcal U(0,1)$, detach the
boundary velocities, and construct
\begin{equation}
  \begin{aligned}
  \bar v_A &= \sg[v_\theta(x_A,0)],
  &\qquad x_t^A &= x_A+t\bar v_A,\\
  \bar v_B &= \sg[v_\theta(x_B,1)],
  &\qquad x_t^B &= x_B-(1-t)\bar v_B.
  \end{aligned}
  \label{eq:selffm-paths}
\end{equation}
The superscripts indicate whether an intermediate state is induced by a source from domain $A$ or domain $B$. Although $x_t^B$ is constructed from the reverse trajectory, both $\bar v_A$ and $\bar v_B$ remain velocity targets in the canonical $A\rightarrow B$ orientation. The selfFM objective is
\begin{equation}
  \begin{aligned}
  \mathcal L_{\mathrm{selfFM}}
  ={}&
  \mathbb E_{x_A,t}
  \left\|v_\theta(x_t^A,t)-\bar v_A\right\|_1
  \\
  &+
  \mathbb E_{x_B,t}
  \left\|v_\theta(x_t^B,t)-\bar v_B\right\|_1 .
  \end{aligned}
  \label{eq:selffm}
\end{equation}
Stop-gradient treats the current boundary velocities and the induced path states as fixed pseudo-targets within this loss. SelfFM therefore updates predictions at sampled intermediate states without simultaneously changing the endpoint displacement through the same objective. The endpoint maps remain trainable through the adversarial, cycle, and representation-path objectives.

\paragraph{Discrete cycle closure.}
A shared continuous velocity field can be traversed in both directions, but the learned one-step Euler maps are not necessarily exact inverses. We denote round-trip reconstructions with a tilde: $\tilde{x}_A$ is obtained by translating $\hat{x}_B$ back to domain $A$, and $\tilde{x}_B$ is obtained by translating $\hat{x}_A$ back to domain $B$. They are defined together with the cycle objective as
\begin{equation}
  \begin{aligned}
  \tilde{x}_A
  &= \hat{x}_B-v_\theta(\hat{x}_B,1),\\
  \tilde{x}_B
  &= \hat{x}_A+v_\theta(\hat{x}_A,0),\\
  \mathcal L_{\mathrm{cyc}}
  &=
  \mathbb E_{x_A}\|\tilde{x}_A-x_A\|_1\\
  &\quad+
  \mathbb E_{x_B}\|\tilde{x}_B-x_B\|_1.
  \end{aligned}
  \label{eq:cycle}
\end{equation}
No stop-gradient is applied in this branch. SelfFM and cycle closure act at complementary levels: selfFM supervises intermediate velocity predictions, whereas cycle closure constrains the composition of the discrete endpoint maps.

\subsection{Representation Path-Velocity Regularization}

SelfFM and cycle closure regularize pixel-space velocity consistency and endpoint composition, but neither directly constrains how perceptual representations evolve along the induced trajectory. A straight interpolation in image space can still produce excessive localized feature changes in representation space. We therefore introduce representation path-velocity regularization (RPV), which measures finite-difference feature changes between neighboring trajectory states.

Let $\phi$ be a frozen spatial encoder that outputs a feature map or patch-token grid. For either translation direction, let $x_s$ denote the source and $\hat{x}_d$ its differentiable generated endpoint in the destination domain. Their induced path is
\begin{equation}
  z_\tau=(1-\tau)x_s+\tau\hat{x}_d,
  \qquad
  \tau\sim\mathcal U(0,1-\Delta).
  \label{eq:rpvpath}
\end{equation}
We extract $F_\tau=\phi(z_\tau)$ and $F_{\tau+\Delta}=\phi(z_{\tau+\Delta})$. At each spatial location $p\in\Omega$, the channel vector is normalized as
\[
  \bar F_{\tau,p}
  =
  \frac{F_{\tau,p}}
       {\|F_{\tau,p}\|_2+\epsilon}.
\]
RPV averages the normalized finite-difference velocity over spatial locations and both directions:
\begin{equation}
  \mathcal L_{\mathrm{RPV}}
  =
  \sum_{\delta\in\{A\rightarrow B,B\rightarrow A\}}
  \mathbb E_{\tau}
  \left[
  \frac{1}{|\Omega|}
  \sum_{p\in\Omega}
  \frac{
  \left\|
  \bar F_{\tau+\Delta,p}^{\,\delta}
  -
  \bar F_{\tau,p}^{\,\delta}
  \right\|_2
  }{\Delta}
  \right].
  \label{eq:rpv}
\end{equation}
The direction index $\delta$ selects the corresponding source--endpoint path. Because features are compared at matching grid locations, the loss remains sensitive to localized changes that may be obscured by globally pooled representations. The encoder parameters are fixed, while gradients propagate through $\phi$ to the generated endpoint and the velocity field.

\subsection{Adversarial Boundary Matching and Overall Objective}

We use one PatchGAN discriminator for each domain to match the generated endpoint marginals to the real data distributions. Let $\bar A=B$ and $\bar B=A$, so that $\hat{x}_d$ denotes an endpoint generated in domain $d$ from a source $x_{\bar d}$. Using least-squares adversarial learning \cite{mao2017lsgan}, the discriminator objective is written compactly as
\begin{equation}
  \mathcal L_D
  =
  \sum_{d\in\{A,B\}}
  \left[
  \mathbb E_{x_d}
  (D_d(x_d)-1)^2
  +
  \mathbb E_{x_{\bar d}}
  D_d(\sg[\hat{x}_d])^2
  \right],
  \label{eq:discriminator}
\end{equation}
and the generator-side adversarial objective is
\begin{equation}
  \mathcal L_{\mathrm{adv}}
  =
  \sum_{d\in\{A,B\}}
  \mathbb E_{x_{\bar d}}
  \left(D_d(\hat{x}_d)-1\right)^2.
  \label{eq:adv}
\end{equation}
Generated endpoints are detached only during discriminator updates. When optimizing the velocity field, gradients from $\mathcal L_{\mathrm{adv}}$ pass through the endpoint maps.

The complete generator objective is
\begin{equation}
  \mathcal L_G
  =
  \lambda_{\mathrm{adv}}\mathcal L_{\mathrm{adv}}
  +
  \lambda_{\mathrm{selfFM}}\mathcal L_{\mathrm{selfFM}}
  +
  \lambda_{\mathrm{cyc}}\mathcal L_{\mathrm{cyc}}
  +
  \lambda_{\mathrm{RPV}}\mathcal L_{\mathrm{RPV}}.
  \label{eq:total}
\end{equation}
The loss weights and representation settings are specified in the implementation details.

\subsection{Inference}

At inference, the discriminators, cycle branch, selfFM sampling, and frozen representation encoder are removed. Translation in either direction follows the Euler updates in Eq.~\eqref{eq:bidirectional-integration}. The main comparison uses $K=1$, for which the outputs are exactly the endpoint predictions in Eq.~\eqref{eq:endpoints}. The same trained velocity field can also be evaluated with a larger integration budget without retraining; we examine $K\in\{1,2,5,10\}$ in the integration-budget ablation.

\section{Experiments}

\begin{table*}[t]
\centering
\begingroup
\fontsize{7.4}{8.6}\selectfont
\setlength{\tabcolsep}{0pt}
\renewcommand{\arraystretch}{1.06}
\begin{tabular*}{\textwidth}{@{\extracolsep{\fill}}l c *{11}{c}@{}}
\toprule
Method & NFE & S$\to$W & W$\to$S & H$\to$Z & Z$\to$H & C$\to$D & D$\to$C & Wi$\to$D & D$\to$Wi & A$\to$M & M$\to$A & Avg. \\
\midrule
CycleGAN & 1 & 81.3 & 77.4 & 68.5 & 152.6 & 158.7 & 166.5 & 136.5 & 95.3 & 382.6 & 56.2 & 137.6 \\
One-to-one CycleGAN & 1 & 85.4 & 81.2 & 71.7 & 156.4 & 134.5 & 142.7 & 123.7 & 96.2 & 356.4 & 59.7 & 130.8 \\
CUT & 1 & 86.1 & 88.9 & 47.2 & 168.2 & 78.5 & 28.4 & 94.7 & 88.2 & 82.6 & 54.3 & 81.7 \\
DCLGAN & 1 & 81.4 & 80.1 & 41.6 & 137.8 & 62.1 & 21.5 & 50.6 & 22.7 & 75.8 & 52.5 & 62.6 \\
LSeSim & 1 & 95.4 & 92.1 & 36.7 & 155.6 & 70.6 & 52.9 & 80.3 & 46.4 & 78.1 & 78.4 & 78.6 \\
MUNIT & 1 & 116.7 & 95.6 & 132.4 & 246.8 & 122.8 & 128.4 & 118.8 & 104.8 & 132.6 & 179.5 & 137.8 \\
UNIT & 1 & 109.4 & 98.2 & 128.9 & 205.6 & 108.7 & 116.2 & 115.2 & 98.6 & 301.5 & 188.4 & 147.1 \\
DistanceGAN & 1 & 95.8 & 102.5 & 69.4 & 168.9 & 118.6 & 121.6 & 108.4 & 92.6 & 329.8 & 96.4 & 130.4 \\
GcGAN & 1 & 99.0 & 104.6 & 84.8 & 175.4 & 146.2 & 154.3 & 129.4 & 85.3 & 223.3 & 81.7 & 128.4 \\
SANTA & 1 & 85.6 & 90.4 & 34.9 & 152.9 & 50.3 & 45.1 & 70.9 & 41.8 & \best{70.3} & 72.4 & 71.5 \\
UNSB & 5 & \best{72.1} & 78.7 & 38.5 & 142.8 & 64.7 & 46.5 & 68.2 & \best{17.3} & 81.4 & 67.4 & 67.8 \\
\midrule
UniCycleFlow & 1 & 81.5 & \second{76.6} & \second{33.1} & \second{116.8} & \second{37.6} & \second{20.9} & \second{37.7} & 20.8 & 74.5 & \second{51.0} & \second{55.1} \\
UniCycleFlow ($K=5$) & 5 & \second{80.9} & \best{75.9} & \best{32.6} & \best{114.5} & \best{36.8} & \best{20.4} & \best{36.9} & \second{20.2} & \second{73.6} & \best{50.2} & \best{54.2} \\
\bottomrule
\end{tabular*}
\endgroup
\caption{Direction-wise FID; lower is better. All displayed baselines are evaluated using our reimplementations with identical data splits and metric code. The best and second-best result in each column are shown in boldface and with an underline, respectively. Published partial results for SDEdit, EGSDE, DDIB, and IBCD are reported in the supplementary material. S/W: Summer/Winter, H/Z: Horse/Zebra, C/D: Cat/Dog, Wi: Wild, and A/M: Aerial/Map.}
\label{tab:main-fid}
\end{table*}

\begin{table*}[t]
\centering
\begingroup
\fontsize{7.4}{8.6}\selectfont
\setlength{\tabcolsep}{0pt}
\renewcommand{\arraystretch}{1.06}
\begin{tabular*}{\textwidth}{@{\extracolsep{\fill}}l c *{11}{c}@{}}
\toprule
Method & NFE & S$\to$W & W$\to$S & H$\to$Z & Z$\to$H & C$\to$D & D$\to$C & Wi$\to$D & D$\to$Wi & A$\to$M & M$\to$A & Avg. \\
\midrule
CycleGAN & 1 & 2.869 & 2.844 & 2.022 & 6.315 & 5.875 & 6.369 & 4.981 & 2.747 & 19.681 & 1.844 & 5.555 \\
One-to-one CycleGAN & 1 & 2.916 & 3.401 & 2.402 & 6.716 & 5.028 & 4.772 & 4.076 & 2.819 & 16.194 & 1.778 & 5.010 \\
CUT & 1 & 3.644 & 3.256 & 1.546 & 7.379 & 3.049 & 0.973 & 3.245 & 3.027 & 3.317 & 1.900 & 3.134 \\
DCLGAN & 1 & 2.995 & 3.325 & 1.456 & 6.028 & 2.430 & \best{0.790} & 1.658 & 0.797 & \best{2.831} & 1.844 & 2.415 \\
LSeSim & 1 & 3.577 & 3.506 & 1.275 & 6.125 & 2.213 & 1.758 & 2.789 & 1.281 & 3.213 & 2.795 & 2.853 \\
MUNIT & 1 & 4.755 & 3.572 & 4.977 & 11.981 & 4.387 & 4.104 & 3.785 & 3.122 & 5.830 & 7.262 & 5.377 \\
UNIT & 1 & 4.291 & 3.288 & 4.863 & 8.735 & 3.407 & 4.300 & 4.646 & 3.090 & 15.803 & 7.899 & 6.032 \\
DistanceGAN & 1 & 3.431 & 3.682 & 2.622 & 7.113 & 3.757 & 4.491 & 3.882 & 2.327 & 15.701 & 3.362 & 5.037 \\
GcGAN & 1 & 3.257 & 3.808 & 3.004 & 7.205 & 4.741 & 5.321 & 4.526 & 2.699 & 10.829 & 3.023 & 4.841 \\
SANTA & 1 & 3.349 & 3.029 & \best{1.105} & 6.770 & 1.691 & 1.512 & 2.297 & 1.347 & \second{3.055} & 2.694 & 2.685 \\
UNSB & 5 & 2.835 & 2.597 & \second{1.220} & 5.718 & 2.389 & 1.384 & 2.280 & \best{0.668} & 3.521 & 2.266 & 2.488 \\
\midrule
UniCycleFlow & 1 & \second{2.820} & \second{2.590} & 1.390 & \second{4.860} & \second{1.345} & 0.895 & \second{1.410} & 0.820 & 3.180 & \second{1.760} & \second{2.107} \\
UniCycleFlow ($K=5$) & 5 & \best{2.749} & \best{2.573} & 1.340 & \best{4.765} & \best{1.299} & \second{0.858} & \best{1.322} & \second{0.783} & 3.113 & \best{1.739} & \best{2.054} \\
\bottomrule
\end{tabular*}
\endgroup
\caption{Direction-wise KID$\times100$; lower is better. All displayed baselines are evaluated using our reimplementations with identical data splits and metric code. The best and second-best result in each column are shown in boldface and with an underline, respectively.}
\label{tab:main-kid}
\end{table*}

\subsection{Experimental Setup}
\label{sec:experimental-setup}

\paragraph{Datasets.}
We evaluate Summer$\leftrightarrow$Winter, Horse$\leftrightarrow$Zebra, Cat$\leftrightarrow$Dog, Wild$\leftrightarrow$Dog, and Aerial$\leftrightarrow$Map~\cite{zhu2017cyclegan,isola2017pix2pix,choi2020starganv2} in both directions, yielding ten translation tasks. For Maps, the two domains are shuffled to ensure pairing information is not used during training. Dataset statistics, resolutions, and preprocessing details are provided in the supplementary material.

\paragraph{Baselines and evaluation.}
We compare with representative cycle-consistent, shared-mapping, contrastive, diffusion, and bridge-based methods. All methods in Tables~\ref{tab:main-fid} and~\ref{tab:main-kid} are evaluated using our implementations with identical splits and metric code; one-way methods are trained separately for each direction. NFE denotes the number of network evaluations and equals $K$ for UniCycleFlow. We report direction-wise Fr\'echet Inception Distance (FID) \cite{heusel2017fid} and Kernel Inception Distance (KID) \cite{binkowski2018kid}; KID is multiplied by $100$. Further details are provided in the supplementary material.

\paragraph{Implementation details.}
We parameterize the velocity field with a four-stage SongUNet of base
width 64 and sinusoidal time conditioning \cite{song2021scorebased},
and use one PatchGAN discriminator per domain. Models are trained for
200 epochs with Adam \cite{kingma2015adam}, batch size 4, and an initial
learning rate of $2\times10^{-4}$. We use frozen VGG16
\texttt{relu5\_3} features~\cite{simonyan2015vgg}, $\Delta=0.1$,
$\lambda_{\mathrm{RPV}}=0.05$, and unit weights for the other objectives.
Main results use NFE=1. Additional optimization, architecture, preprocessing, and evaluation details are provided in the supplementary material.

\begin{figure*}[t]
\centering
\includegraphics[
    width=\linewidth,
    height=0.60\textheight,
    keepaspectratio
]{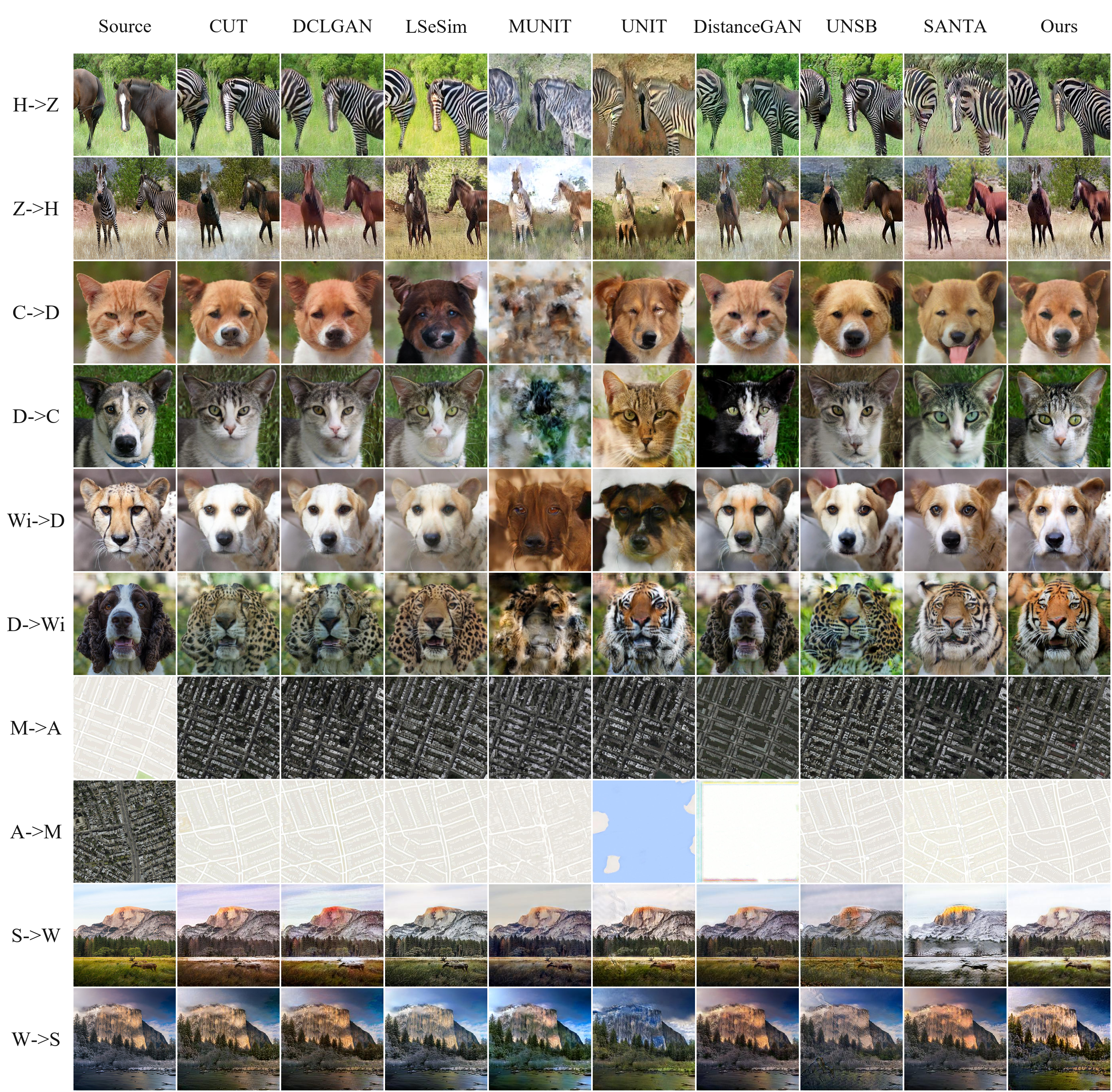}
\caption{Qualitative comparison across the five domain pairs using identical source images. UniCycleFlow produces visually plausible target-domain translations in both directions while retaining the major spatial structure of each source image.}
\label{fig:qualitative}
\end{figure*}

\subsection{Comparison with State of the Art}
Tables~\ref{tab:main-fid} and~\ref{tab:main-kid} report results across ten translation directions. With a single Euler evaluation, UniCycleFlow achieves the lowest FID on 7 directions and the best ten-direction average FID and KID$\times100$ of 55.1 and 2.107. Relative to DCLGAN, the strongest prior ten-direction average, this reduces FID from 62.6 to 55.1 and KID$\times100$ from 2.415 to 2.107. With five evaluations, the same field further improves the averages to 54.2 and 2.054.

Figure~\ref{fig:qualitative} compares both directions using identical source images. UniCycleFlow produces plausible target-domain translations across all five domain pairs; additional results are provided in the supplementary material.

\subsection{Ablation Studies}
\label{sec:ablation}
For compact presentation, the main-paper ablations report Horse$\leftrightarrow$Zebra together with FID and KID averaged over all ten translation directions. Complete direction-wise results are provided in the supplementary material. Unless otherwise specified, the training-objective, field-organization, and representation ablations are evaluated with NFE=5; the inference-budget study separately reports NFE=1, 2, 5, and 10.

\paragraph{Component contributions.}
Table~\ref{tab:core-ablation} evaluates how the proposed components progressively improve the learned transport. Independent-pair rectified-flow training yields an average FID of 113.5, as arbitrary cross-domain pairings introduce substantial instance-level variation into the velocity targets. Learning source-conditioned endpoints through adversarial boundary matching provides the largest improvement, reducing the average FID to 69.9. SelfFM and cycle closure further improve the results individually, and their combination reaches average FID and KID$\times100$ of 55.9 and 2.146, respectively. Adding RPV improves the averages to 54.2 and 2.054 without changing the inference graph. The progressive gains are consistent with the complementary roles of source-conditioned coupling, intermediate velocity supervision, discrete cycle closure, and representation-space path regularization.

\begin{table}[t]
\centering
\small
\setlength{\tabcolsep}{2.5pt}
\begin{tabular}{@{}lrrrr@{}}
\toprule
Setting & H$\to$Z & Z$\to$H & Avg. FID & Avg. KID$\times100$ \\
\midrule
Independent-pair RF & 111.8 & 186.3 & 113.5 & 5.159 \\
Adversarial endpoints & 55.2 & 143.1 & 69.9 & 2.981 \\
+ selfFM & 43.7 & 133.4 & 62.0 & 2.457 \\
+ Cycle & 45.6 & 129.0 & 61.7 & 2.443 \\
+ selfFM + Cycle
& \best{32.4} & \second{118.8}
& \second{55.9} & \second{2.146} \\
+ RPV (full)
& \second{32.6} & \best{114.5}
& \best{54.2} & \best{2.054} \\
\bottomrule
\end{tabular}
\caption{Component ablation. All rows are evaluated with NFE=5.}
\label{tab:core-ablation}
\end{table}

\paragraph{Velocity-field organization.}
Table~\ref{tab:field-organization} examines the effects of parameter sharing and temporal organization. Two independently parameterized fields obtain an average FID of 76.2. Sharing the backbone while retaining direction-specific output heads improves the average FID to 60.5, showing that the two directions benefit from common feature representations. A shared field conditioned by a discrete direction token reaches an average FID of 62.2, whereas organizing the shared field continuously over time further improves the average FID to 54.2. The KID results exhibit the same overall trend. These comparisons support the proposed formulation in which both directions share not only network parameters, but also a common temporal organization of cross-domain change.

\begin{table}[t]
\centering
\small
\setlength{\tabcolsep}{1.7pt}
\renewcommand{\arraystretch}{1.05}
\begin{tabular}{@{}lrrrr@{}}
\toprule
Field design
& H$\to$Z
& Z$\to$H
& \shortstack{Avg.\\FID}
& \shortstack{Avg. KID\\$\times100$} \\
\midrule
Separate fields
& 42.2 & 142.9 & 76.2 & 3.065 \\
Shared backbone + two heads
& 32.8 & 116.1 & 60.5 & 2.336 \\
Shared field + direction token
& 33.7 & 118.2 & 62.2 & 2.433 \\
Shared time-conditioned field
& \best{32.6} & \best{114.5}
& \best{54.2} & \best{2.054} \\
\bottomrule
\end{tabular}
\caption{Velocity-field organization ablation. All variants use the same training objectives and are evaluated with NFE=5.}
\label{tab:field-organization}
\end{table}

\paragraph{Inference-time integration budget.}
Table~\ref{tab:nfe} evaluates the same trained velocity field using different Euler step counts. With a single evaluation, the model already reaches average FID and KID$\times100$ of 55.1 and 2.107, which are close to the best measured values. Increasing the budget to two evaluations improves the averages to 54.3 and 2.059, while five evaluations yield 54.2 and 2.054. The relatively small gap between one-step and multi-step inference shows that the boundary prediction provides a strong direct translation, while intermediate field evaluations offer a modest additional refinement. This behavior is consistent with the role of selfFM and RPV in regularizing states along the model-induced path.

\begin{table}[t]
\centering
\small
\setlength{\tabcolsep}{3.0pt}
\begin{tabular}{@{}crrrr@{}}
\toprule
NFE & H$\to$Z & Z$\to$H & Avg. FID & Avg. KID$\times100$ \\
\midrule
1 & 33.1 & 116.8 & 55.1 & 2.107 \\
2 & \best{32.4} & \second{115.3}
& \second{54.3} & \second{2.059} \\
5 & \second{32.6} & \best{114.5}
& \best{54.2} & \best{2.054} \\
10 & 35.8 & 121.7 & 57.3 & 2.228 \\
\bottomrule
\end{tabular}
\caption{Euler-step ablation using the same trained velocity field.}
\label{tab:nfe}
\end{table}

\paragraph{RPV design.}
Complete encoder and hyperparameter studies are reported in the
supplementary material. VGG16 \texttt{relu5\_3} with $\Delta=0.1$
and $\lambda_{\mathrm{RPV}}=0.05$ gives the best overall averages,
while DINO and DINOv2 representations
\cite{caron2021dino,oquab2023dinov2} and nearby parameter choices
remain competitive.

\section{Discussion and Limitations}

UniCycleFlow uses a deterministic shared velocity field and therefore
does not explicitly model one-to-many outputs. Although this design
supports efficient and structure-preserving translation, a stochastic
style variable or a distribution over source-conditioned endpoints
could improve output diversity without abandoning bidirectional
sharing. SelfFM relies on detached boundary predictions as
pseudo-targets, so inaccurate endpoints, especially early in training,
may bias intermediate supervision. Teacher-based or
confidence-weighted targets could reduce this dependence.

RPV further depends on the geometry of the frozen representation.
The encoder ablation shows that VGG16 performs best on average, while
DINO-based features remain competitive, suggesting that both spatial
correspondence and semantic invariance matter. Moreover, the current
objective mainly constrains the magnitude of local feature changes,
rather than their semantic direction. Target-aware feature directions
or adaptive multi-scale representations could provide more precise
trajectory supervision.

\section{Conclusion}

We presented UniCycleFlow, which realizes bidirectional unpaired translation by traversing a single time-conditioned velocity field in opposite directions. Source-conditioned adversarial endpoints construct translation-specific paths, while selfFM, cycle closure, and RPV regularize intermediate dynamics, discrete consistency, and representation-space changes. UniCycleFlow achieves the best average FID and KID across ten directions with one-step inference.

\clearpage
\raggedbottom
\setlength{\bibsep}{0pt}
\bibliography{UniCycleFlow_AAAI2027}

@inproceedings{isola2017pix2pix,
  title={Image-to-Image Translation with Conditional Adversarial Networks},
  author={Isola, Phillip and Zhu, Jun-Yan and Zhou, Tinghui and Efros, Alexei A.},
  booktitle={Proceedings of the IEEE Conference on Computer Vision and Pattern Recognition},
  pages={1125--1134},
  year={2017}
}

@inproceedings{zhu2017cyclegan,
  title={Unpaired Image-to-Image Translation Using Cycle-Consistent Adversarial Networks},
  author={Zhu, Jun-Yan and Park, Taesung and Isola, Phillip and Efros, Alexei A.},
  booktitle={Proceedings of the IEEE International Conference on Computer Vision},
  pages={2223--2232},
  year={2017}
}

@inproceedings{yi2017dualgan,
  title={DualGAN: Unsupervised Dual Learning for Image-to-Image Translation},
  author={Yi, Zili and Zhang, Hao and Tan, Ping and Gong, Minglun},
  booktitle={Proceedings of the IEEE International Conference on Computer Vision},
  pages={2849--2857},
  year={2017}
}

@inproceedings{kim2017discogan,
  title={Learning to Discover Cross-Domain Relations with Generative Adversarial Networks},
  author={Kim, Taeksoo and Cha, Moonsu and Kim, Hyunsoo and Lee, Jung Kwon and Kim, Jiwon},
  booktitle={Proceedings of the 34th International Conference on Machine Learning},
  pages={1857--1865},
  year={2017}
}

@inproceedings{liu2017unit,
  title={Unsupervised Image-to-Image Translation Networks},
  author={Liu, Ming-Yu and Breuel, Thomas and Kautz, Jan},
  booktitle={Advances in Neural Information Processing Systems},
  year={2017}
}

@inproceedings{huang2018munit,
  title={Multimodal Unsupervised Image-to-Image Translation},
  author={Huang, Xun and Liu, Ming-Yu and Belongie, Serge and Kautz, Jan},
  booktitle={Proceedings of the European Conference on Computer Vision},
  pages={172--189},
  year={2018}
}

@inproceedings{choi2018stargan,
  title={StarGAN: Unified Generative Adversarial Networks for Multi-Domain Image-to-Image Translation},
  author={Choi, Yunjey and Choi, Minje and Kim, Munyoung and Ha, Jung-Woo and Kim, Sunghun and Choo, Jaegul},
  booktitle={Proceedings of the IEEE Conference on Computer Vision and Pattern Recognition},
  pages={8789--8797},
  year={2018}
}

@inproceedings{choi2020starganv2,
  title={StarGAN v2: Diverse Image Synthesis for Multiple Domains},
  author={Choi, Yunjey and Uh, Youngjung and Yoo, Jaejun and Ha, Jung-Woo},
  booktitle={Proceedings of the IEEE/CVF Conference on Computer Vision and Pattern Recognition},
  pages={8188--8197},
  year={2020}
}

@inproceedings{shen2020one,
  title={One-to-One Mapping for Unpaired Image-to-Image Translation},
  author={Shen, Zengming and Zhou, S. Kevin and Chen, Yifan and Georgescu, Bogdan and Liu, Xuqi and Huang, Thomas S.},
  booktitle={Proceedings of the IEEE Winter Conference on Applications of Computer Vision},
  pages={1170--1179},
  year={2020}
}

@inproceedings{park2020cut,
  title={Contrastive Learning for Unpaired Image-to-Image Translation},
  author={Park, Taesung and Efros, Alexei A. and Zhang, Richard and Zhu, Jun-Yan},
  booktitle={Proceedings of the European Conference on Computer Vision},
  pages={319--345},
  year={2020}
}

@inproceedings{han2021dclgan,
  title={Dual Contrastive Learning for Unsupervised Image-to-Image Translation},
  author={Han, Junlin and Shoeiby, Mehrdad and Petersson, Lars and Armin, Mohammad Ali},
  booktitle={Proceedings of the IEEE/CVF Conference on Computer Vision and Pattern Recognition Workshops},
  pages={746--755},
  year={2021}
}

@inproceedings{zheng2021spatial,
  title={The Spatially-Correlative Loss for Various Image Translation Tasks},
  author={Zheng, Chuanxia and Cham, Tat-Jen and Cai, Jianfei},
  booktitle={Proceedings of the IEEE/CVF Conference on Computer Vision and Pattern Recognition},
  pages={16407--16417},
  year={2021}
}

@inproceedings{xie2023shortest,
  title={Unpaired Image-to-Image Translation with Shortest Path Regularization},
  author={Xie, Shaoan and Xu, Yanwu and Gong, Mingming and Zhang, Kun},
  booktitle={Proceedings of the IEEE/CVF Conference on Computer Vision and Pattern Recognition},
  pages={10177--10187},
  year={2023}
}

@inproceedings{torbunov2023uvcgan,
  title={UVCGAN: UNet Vision Transformer Cycle-Consistent GAN for Unpaired Image-to-Image Translation},
  author={Torbunov, Dmitrii and Huang, Yi and Yu, Haiwang and Huang, Jin and Yoo, Shinjae and Lin, Meifeng and Viren, Brett and Ren, Yihui},
  booktitle={Proceedings of the IEEE/CVF Winter Conference on Applications of Computer Vision},
  pages={702--712},
  year={2023}
}

@misc{torbunov2023uvcganv2,
  title={UVCGAN v2: An Improved Cycle-Consistent GAN for Unpaired Image-to-Image Translation},
  author={Torbunov, Dmitrii and Huang, Yi and Tseng, Huan-Hsin and Yu, Haiwang and Huang, Jin and Yoo, Shinjae and Lin, Meifeng and Viren, Brett and Ren, Yihui},
  year={2023},
  eprint={2303.16280},
  archivePrefix={arXiv}
}

@inproceedings{lipman2023flow,
  title={Flow Matching for Generative Modeling},
  author={Lipman, Yaron and Chen, Ricky T. Q. and Ben-Hamu, Heli and Nickel, Maximilian and Le, Matt},
  booktitle={International Conference on Learning Representations},
  year={2023}
}

@inproceedings{liu2023rectified,
  title={Flow Straight and Fast: Learning to Generate and Transfer Data with Rectified Flow},
  author={Liu, Xingchao and Gong, Chengyue and Liu, Qiang},
  booktitle={International Conference on Learning Representations},
  year={2023}
}

@inproceedings{su2023ddib,
  title={Dual Diffusion Implicit Bridges for Image-to-Image Translation},
  author={Su, Xuan and Song, Jiaming and Meng, Chenlin and Ermon, Stefano},
  booktitle={International Conference on Learning Representations},
  year={2023}
}

@inproceedings{kim2024unsb,
  title={Unpaired Image-to-Image Translation via Neural Schr\"odinger Bridge},
  author={Kim, Beomsu and Kwon, Gihyun and Kim, Kwanyoung and Ye, Jong Chul},
  booktitle={International Conference on Learning Representations},
  year={2024}
}

@misc{lee2025ibcd,
  title={Single-Step Bidirectional Unpaired Image Translation Using Implicit Bridge Consistency Distillation},
  author={Lee, Suhyeon and Kim, Kwanyoung and Ye, Jong Chul},
  year={2025},
  eprint={2503.15056},
  archivePrefix={arXiv}
}

@inproceedings{mao2017lsgan,
  title={Least Squares Generative Adversarial Networks},
  author={Mao, Xudong and Li, Qing and Xie, Haoran and Lau, Raymond Y. K. and Wang, Zhen and Smolley, Stephen Paul},
  booktitle={Proceedings of the IEEE International Conference on Computer Vision},
  pages={2794--2802},
  year={2017}
}

@inproceedings{simonyan2015vgg,
  title={Very Deep Convolutional Networks for Large-Scale Image Recognition},
  author={Simonyan, Karen and Zisserman, Andrew},
  booktitle={International Conference on Learning Representations},
  year={2015}
}

@inproceedings{heusel2017fid,
  title={GANs Trained by a Two Time-Scale Update Rule Converge to a Local Nash Equilibrium},
  author={Heusel, Martin and Ramsauer, Hubert and Unterthiner, Thomas and Nessler, Bernhard and Hochreiter, Sepp},
  booktitle={Advances in Neural Information Processing Systems},
  year={2017}
}

@inproceedings{binkowski2018kid,
  title={Demystifying MMD GANs},
  author={Bi{\'n}kowski, Miko{\l}aj and Sutherland, Dougal J. and Arbel, Michael and Gretton, Arthur},
  booktitle={International Conference on Learning Representations},
  year={2018}
}

@inproceedings{meng2022sdedit,
  title={SDEdit: Guided Image Synthesis and Editing with Stochastic Differential Equations},
  author={Meng, Chenlin and He, Yutong and Song, Yang and Song, Jiaming and Wu, Jiajun and Zhu, Jun-Yan and Ermon, Stefano},
  booktitle={International Conference on Learning Representations},
  year={2022}
}

@inproceedings{zhao2022egsde,
  title={EGSDE: Unpaired Image-to-Image Translation via Energy-Guided Stochastic Differential Equations},
  author={Zhao, Min and Bao, Fan and Li, Chongxuan and Zhu, Jun},
  booktitle={Advances in Neural Information Processing Systems},
  year={2022}
}

@inproceedings{caron2021dino,
  title={Emerging Properties in Self-Supervised Vision Transformers},
  author={Caron, Mathilde and Touvron, Hugo and Misra, Ishan and J\'egou, Herv\'e and Mairal, Julien and Bojanowski, Piotr and Joulin, Armand},
  booktitle={Proceedings of the IEEE/CVF International Conference on Computer Vision},
  pages={9650--9660},
  year={2021}
}

@article{oquab2023dinov2,
  title={DINOv2: Learning Robust Visual Features without Supervision},
  author={Oquab, Maxime and Darcet, Timoth\'ee and Moutakanni, Th\'eo and Vo, Huy and Szafraniec, Marc and Khalidov, Vasil and Fernandez, Pierre and Haziza, Daniel and Massa, Francisco and El-Nouby, Alaaeldin and others},
  journal={Transactions on Machine Learning Research},
  year={2024}
}

@inproceedings{kingma2015adam,
  title={Adam: A Method for Stochastic Optimization},
  author={Kingma, Diederik P. and Ba, Jimmy},
  booktitle={International Conference on Learning Representations},
  year={2015}
}

@inproceedings{zhang2018lpips,
  title={The Unreasonable Effectiveness of Deep Features as a Perceptual Metric},
  author={Zhang, Richard and Isola, Phillip and Efros, Alexei A. and Shechtman, Eli and Wang, Oliver},
  booktitle={Proceedings of the IEEE/CVF Conference on Computer Vision and Pattern Recognition},
  pages={586--595},
  year={2018}
}

@inproceedings{song2023shunit,
  title     = {{SHUNIT}: Style Harmonization for Unpaired Image-to-Image Translation},
  author    = {Song, Seokbeom and Lee, Suhyeon and Seong, Hongje and Min, Kyoungwon and Kim, Euntai},
  booktitle = {Proceedings of the AAAI Conference on Artificial Intelligence},
  volume    = {37},
  number    = {2},
  pages     = {2292--2302},
  year      = {2023},
  doi       = {10.1609/aaai.v37i2.25324}
}

@inproceedings{cai2024enco,
  title     = {Rethinking the Paradigm of Content Constraints in Unpaired Image-to-Image Translation},
  author    = {Cai, Xiuding and Zhu, Yaoyao and Miao, Dong and Fu, Linjie and Yao, Yu},
  booktitle = {Proceedings of the AAAI Conference on Artificial Intelligence},
  volume    = {38},
  number    = {2},
  pages     = {891--899},
  year      = {2024},
  doi       = {10.1609/aaai.v38i2.27848}
}

@inproceedings{wu2024stegogan,
  title     = {{StegoGAN}: Leveraging Steganography for Non-Bijective Image-to-Image Translation},
  author    = {Wu, Sidi and Chen, Yizi and Mermet, Samuel and Hurni, Lorenz and Schindler, Konrad and Gonthier, Nicolas and Landrieu, Loic},
  booktitle = {Proceedings of the IEEE/CVF Conference on Computer Vision and Pattern Recognition},
  pages     = {7922--7931},
  year      = {2024}
}

@inproceedings{song2021scorebased,
  title     = {Score-Based Generative Modeling through Stochastic Differential Equations},
  author    = {Song, Yang and Sohl-Dickstein, Jascha and Kingma, Diederik P. and Kumar, Abhishek and Ermon, Stefano and Poole, Ben},
  booktitle = {International Conference on Learning Representations},
  year      = {2021}
}

@article{benaim2017one,
  title={One-sided unsupervised domain mapping},
  author={Benaim, Sagie and Wolf, Lior},
  journal={Advances in neural information processing systems},
  volume={30},
  year={2017}
}

@inproceedings{fu2019geometry,
  title={Geometry-consistent generative adversarial networks for one-sided unsupervised domain mapping},
  author={Fu, Huan and Gong, Mingming and Wang, Chaohui and Batmanghelich, Kayhan and Zhang, Kun and Tao, Dacheng},
  booktitle={Proceedings of the IEEE/CVF conference on computer vision and pattern recognition},
  pages={2427--2436},
  year={2019}
}

@article{xie2022density,
  title={Unsupervised image-to-image translation with density changing regularization},
  author={Xie, Shaoan and Ho, Qirong and Zhang, Kun},
  journal={Advances in Neural Information Processing Systems},
  volume={35},
  pages={28545--28558},
  year={2022}
}

\clearpage
\twocolumn[{
\centering
{\Large\bfseries Supplementary Material\par}
\vspace{3pt}
{\large
UniCycleFlow: Bidirectional Unpaired Image Translation with a Shared Rectified Flow
\par}
\vspace{6pt}
}]

\setcounter{topnumber}{5}
\setcounter{bottomnumber}{2}
\setcounter{totalnumber}{7}
\setcounter{dbltopnumber}{5}
\renewcommand{\topfraction}{0.95}
\renewcommand{\bottomfraction}{0.85}
\renewcommand{\textfraction}{0.05}
\renewcommand{\floatpagefraction}{0.85}
\renewcommand{\dbltopfraction}{0.95}
\renewcommand{\dblfloatpagefraction}{0.85}
\setlength{\textfloatsep}{8pt plus 1pt minus 2pt}
\setlength{\floatsep}{7pt plus 1pt minus 1pt}
\setlength{\intextsep}{8pt plus 1pt minus 2pt}

This supplementary appendix provides the experimental protocol, additional published FID comparisons, additional qualitative results, and full cross-dataset ablations supporting the main paper. We first document the datasets, preprocessing, optimization, evaluation protocol, and organization of compared methods. We then report the expanded FID comparison, followed by additional qualitative comparisons and all ten-direction ablations for the training objectives, velocity-field organization, inference budget, and RPV design.

\section{Experimental Protocol}
\label{sec:supp-protocol}

\subsection{Datasets and Preprocessing}
We evaluate five domain pairs in both directions, yielding ten translation tasks. Table~\ref{tab:datasets} summarizes the split sizes and native resolutions. A/B follow the domain order in each benchmark name. Summer$\leftrightarrow$Winter and Horse$\leftrightarrow$Zebra use the CycleGAN dataset release \cite{zhu2017cyclegan}, Aerial$\leftrightarrow$Map uses the pix2pix/CycleGAN Maps release \cite{zhu2017cyclegan}, and Cat$\leftrightarrow$Dog and Wild$\leftrightarrow$Dog are constructed from the corresponding AFHQ domains \cite{choi2020starganv2}. All reimplemented methods use the same split files and the same dataset-specific preprocessing pipeline.

\begin{table}[!t]
\centering
\small
\begin{tabular}{lccc}
\toprule
Benchmark & Train A/B & Test A/B & Native Res. \\
\midrule
Summer$\leftrightarrow$Winter & 1231/962 & 309/238 & $256^2$ \\
Horse$\leftrightarrow$Zebra & 1067/1334 & 120/140 & $256^2$ \\
Cat$\leftrightarrow$Dog & 5153/4739 & 500/500 & $512^2$ \\
Wild$\leftrightarrow$Dog & 4738/4739 & 500/500 & $512^2$ \\
Aerial$\leftrightarrow$Map & 1096/1096 & 1098/1098 & $600^2$ \\
\bottomrule
\end{tabular}
\caption{Dataset split sizes and native resolutions for the five evaluated domain pairs.}
\label{tab:datasets}
\end{table}

\paragraph{Training preprocessing.}
For Summer$\leftrightarrow$Winter and Horse$\leftrightarrow$Zebra, images are resized to $286\times286$ using bicubic interpolation, randomly cropped to $256\times256$, and horizontally flipped with probability $0.5$. For Cat$\leftrightarrow$Dog and Wild$\leftrightarrow$Dog, a random resized crop to $256\times256$ is applied with probability $0.5$, using scale range $[0.8,1.0]$ and aspect-ratio range $[0.9,1.1]$; images are then resized to $256\times256$ and horizontally flipped with probability $0.5$. For Aerial$\leftrightarrow$Map, images from the two domains are resized to $512\times512$ using bilinear interpolation, without random cropping or horizontal flipping. The domains are shuffled independently, and no pairing information is used during training.

\paragraph{Evaluation preprocessing.}
Summer$\leftrightarrow$Winter, Horse$\leftrightarrow$Zebra, Cat$\leftrightarrow$Dog, and Wild$\leftrightarrow$Dog use the unaligned evaluation loader and are resized to $256\times256$ using bilinear interpolation. Aerial$\leftrightarrow$Map uses the aligned evaluation loader and is resized to $512\times512$ using bilinear interpolation. No random augmentation is applied during evaluation.

\subsection{Architecture Details}

\paragraph{Velocity network.}
The velocity field is parameterized by a four-stage SongUNet \cite{song2021scorebased} with three input channels and three output channels. The base width is 64, with stage widths $[64,128,128,128]$ and channel multipliers $[1,2,2,2]$. Each encoder stage contains two residual blocks, each decoder stage contains three residual blocks, and the bottleneck contains two residual blocks. Downsampling and upsampling use learned convolutional resampling inside the UNet blocks with resampling filter $[1,1]$. All residual blocks use GroupNorm, SiLU activations, and dropout rate $0.1$. Attention is applied at the bottleneck and at the configured $16\times16$ resolution.

\paragraph{Time conditioning.}
For $t\in[0,1]$, the scalar time is multiplied by $999$ before sinusoidal positional embedding. The sinusoidal embedding has dimension 64 and is processed by a two-layer MLP with hidden and output dimension 256. The resulting embedding is injected into every residual block through an affine projection and added to the normalized feature activations; adaptive scale--shift conditioning is disabled.

\paragraph{Discriminators.}
We use one independently parameterized PatchGAN discriminator for each domain. Each discriminator follows the NLayerDiscriminator design with channels $[64,128,256,512,1]$, kernel size 4, and strides $[2,2,2,1,1]$. The hidden layers use InstanceNorm and LeakyReLU with negative slope $0.2$, without spectral normalization. The output patch map is $30\times30$ for $256\times256$ inputs and $62\times62$ for $512\times512$ inputs.

\subsection{Optimization Details}

UniCycleFlow and its ablations are trained on Rocky Linux 9.7 using a single NVIDIA A100 80GB GPU. We train for 200 epochs with batch size 4 using Adam \cite{kingma2015adam}, an initial learning rate of $2\times10^{-4}$, $\beta_1=0.5$, $\beta_2=0.999$, and zero weight decay. The learning rate is held constant for the first 100 epochs and then linearly decayed to zero. Network weights are initialized with Xavier initialization using gain $0.02$.

The generator and discriminators are updated at a $1{:}1$ ratio, with the generator updated before the discriminators in each iteration. A replay buffer of 50 generated images is used for discriminator training. Training uses BF16 mixed precision, without gradient clipping or gradient accumulation. We use random seed 42. Each reported result is obtained from a single run using the latest training checkpoint, without validation-set model selection.

\subsection{Loss and Inference Details}

The adversarial, selfFM, and cycle losses use unit weights, while $\lambda_{\mathrm{RPV}}=0.05$. The least-squares discriminator loss \cite{mao2017lsgan} applies the conventional factor of $1/2$ to the sum of its real and fake terms; no corresponding factor is applied to the generator-side adversarial loss. Pixel-space losses are computed in the model range $[-1,1]$.

For selfFM, one time value is sampled independently for each image and direction from the uniform distribution on $[0,1)$. For RPV, one path position is likewise sampled independently for each image and direction. The frozen VGG16 encoder \cite{simonyan2015vgg} uses \texttt{relu5\_3} features and LPIPS-style input scaling \cite{zhang2018lpips} rather than ImageNet mean--standard-deviation normalization. The model-range input is passed directly to this scaling layer without first converting it to $[0,1]$. The selected feature grid is $16\times16$ for $256\times256$ inputs and $32\times32$ for $512\times512$ inputs. Feature normalization uses $\epsilon=10^{-10}$, and the speed denominator is clamped at $10^{-8}$. RPV computes the channel-wise $\ell_2$ speed at each spatial location and averages over spatial positions, selected layers, the batch, and both translation directions.

Inference uses explicit Euler integration on the fixed grid $t_k=k/K$. The main comparison uses $K=1$, and the same checkpoint is evaluated with $K\in\{2,5,10\}$ by changing only the number of inference steps. Intermediate Euler states are not clipped; the final output is clipped to $[-1,1]$ and converted to uint8 for saving. Inference uses batch size 1.

For the DINO ablations, we use the publicly released
pretrained DINO ViT-S/16 and DINOv2 ViT-B/14 weights \cite{caron2021dino,oquab2023dinov2}.
The class token is discarded, and the remaining patch tokens
are reshaped to their native spatial grid. Inputs use the
normalization associated with the corresponding pretrained
model, and positional embeddings are interpolated when
required.

\subsection{Evaluation Protocol}

FID \cite{heusel2017fid} and KID \cite{binkowski2018kid} are computed with a custom wrapper around the \texttt{pytorch-fid} InceptionV3 implementation using the TensorFlow/FID Inception weights and the 2048-dimensional pool3 features. For each direction, all target-domain test images form the real set, and one translated image is generated for every source-domain test image. No samples are repeated when a test set is small. All reproduced methods use the same source test images.

KID uses 100 subsets, with each subset size set to
$\min(1000,n_{\mathrm{real}},n_{\mathrm{generated}})$. Metric inputs are saved as uint8 images in $[0,255]$, converted to $[0,1]$, and internally resized to $299\times299$ by the Inception implementation. KID is multiplied by $100$ for reporting. Ten-direction averages are unweighted arithmetic means.

\subsection{Baseline Implementations}

All reproduced baselines are trained for 200 epochs with the dataset-specific image sizes and preprocessing described above, one random seed, and latest-checkpoint evaluation. One-direction methods are trained independently for each direction. We use the same framework data pipeline for all reproduced methods and do not perform dataset-specific hyperparameter retuning unless explicitly stated. Multimodal baselines produce one output per source image, and metrics are computed from that single output rather than averaged over multiple samples.

For each reproduced baseline, we follow the official
architecture and default loss configuration whenever available.
Only the input resolution, dataset loader, training length, and
evaluation pipeline are standardized. 

\subsection{Directionality and Sampling Taxonomy}
Table~\ref{tab:taxonomy} summarizes how each method supports the two translation directions, how its generator is organized, and the NFE used by the reported configuration. The comparison covers cycle-consistent and shared mappings \cite{zhu2017cyclegan,shen2020one}, contrastive and latent-space translation \cite{park2020cut,han2021dclgan,zheng2021spatial,huang2018munit,liu2017unit}, shortest-path regularization \cite{xie2023shortest}, and diffusion or bridge transport \cite{meng2022sdedit,zhao2022egsde,su2023ddib,kim2024unsb,lee2025ibcd}. Direction-specific methods require a separate training run for each direction. Two-generator systems learn both directions jointly but retain direction-specific generator parameters. One-generator systems share the same generator across directions.

\begin{table*}[!t]
\centering
\small
\begin{tabular}{lccc}
\toprule
Method & Direction support & Generator organization & NFE per translation \\
\midrule
CycleGAN & Bidirectional & Two generators & 1 \\
One-to-one CycleGAN & Bidirectional & One generator & 1 \\
CUT & One-way per run & Separate model per direction & 1 \\
DCLGAN & Bidirectional & Two generators & 1 \\
LSeSim & One-way per run & Separate model per direction & 1 \\
MUNIT & Bidirectional & Two generators & 1 \\
UNIT & Bidirectional & Two generators & 1 \\
DistanceGAN & One-way per run & Separate model per direction & 1 \\
GcGAN & One-way per run & Separate model per direction & 1 \\
SANTA & One-way per run & Separate model per direction & 1 \\
SDEdit & One-way per run & Domain diffusion prior & 1000 \\
EGSDE & One-way per run & Domain diffusion prior & 1200 \\
DDIB & Bidirectional & Two domain diffusion models & 160 \\
UNSB & One-way per run & Separate bridge per direction & 5 \\
IBCD & Bidirectional & One distilled bridge model & 1 \\
\method & Bidirectional & One shared velocity network & 1 (main) / 5 (additional) \\
\bottomrule
\end{tabular}
\caption{Directionality, generator organization, and sampling NFE for the configurations reported in the quantitative comparison.}
\label{tab:taxonomy}
\end{table*}

\section{Additional Quantitative Comparison}
\label{sec:supp-main-results}

Table~\ref{tab:supp-main-fid} extends the main-paper FID comparison with partial values quoted directly from prior diffusion and bridge papers without conversion or re-evaluation. The SDEdit and EGSDE values follow the comparison reported in \cite{lee2025ibcd}, whose source study is EGSDE \cite{zhao2022egsde}; the original SDEdit method is described in \cite{meng2022sdedit}. The DDIB (Teacher) and IBCD values are taken from \cite{lee2025ibcd}, with DDIB introduced in \cite{su2023ddib}. Because these methods cover only a subset of directions and may use different dataset, resolution, or FID protocols, their values are provided for reference and excluded from all column-wise rankings. The direction-wise KID table is not repeated here because it is already reported unchanged in Table~\ref{tab:main-kid}.

\begin{table*}[!t]
\centering
\begingroup
\fontsize{7.4}{8.6}\selectfont
\setlength{\tabcolsep}{0pt}
\renewcommand{\arraystretch}{1.06}
\begin{tabular*}{\textwidth}{@{\extracolsep{\fill}}l c *{11}{c}@{}}
\toprule
Method & NFE & S$\to$W & W$\to$S & H$\to$Z & Z$\to$H & C$\to$D & D$\to$C & Wi$\to$D & D$\to$Wi & A$\to$M & M$\to$A & Avg. \\
\midrule
CycleGAN & 1 & 81.3 & 77.4 & 68.5 & 152.6 & 158.7 & 166.5 & 136.5 & 95.3 & 382.6 & 56.2 & 137.6 \\
One-to-one CycleGAN & 1 & 85.4 & 81.2 & 71.7 & 156.4 & 134.5 & 142.7 & 123.7 & 96.2 & 356.4 & 59.7 & 130.8 \\
CUT & 1 & 86.1 & 88.9 & 47.2 & 168.2 & 78.5 & 28.4 & 94.7 & 88.2 & 82.6 & 54.3 & 81.7 \\
DCLGAN & 1 & 81.4 & 80.1 & 41.6 & 137.8 & 62.1 & 21.5 & 50.6 & 22.7 & 75.8 & 52.5 & 62.6 \\
LSeSim & 1 & 95.4 & 92.1 & 36.7 & 155.6 & 70.6 & 52.9 & 80.3 & 46.4 & 78.1 & 78.4 & 78.6 \\
MUNIT & 1 & 116.7 & 95.6 & 132.4 & 246.8 & 122.8 & 128.4 & 118.8 & 104.8 & 132.6 & 179.5 & 137.8 \\
UNIT & 1 & 109.4 & 98.2 & 128.9 & 205.6 & 108.7 & 116.2 & 115.2 & 98.6 & 301.5 & 188.4 & 147.1 \\
DistanceGAN & 1 & 95.8 & 102.5 & 69.4 & 168.9 & 118.6 & 121.6 & 108.4 & 92.6 & 329.8 & 96.4 & 130.4 \\
GcGAN & 1 & 99.0 & 104.6 & 84.8 & 175.4 & 146.2 & 154.3 & 129.4 & 85.3 & 223.3 & 81.7 & 128.4 \\
SANTA & 1 & 85.6 & 90.4 & 34.9 & 152.9 & 50.3 & 45.1 & 70.9 & 41.8 & \best{70.3} & 72.4 & 71.5 \\
UNSB & 5 & \best{72.1} & 78.7 & 38.5 & 142.8 & 64.7 & 46.5 & 68.2 & \best{17.3} & 81.4 & 67.4 & 67.8 \\
\midrule
\ext{SDEdit} & \ext{1000} & \ext{--} & \ext{--} & \ext{--} & \ext{--} & \ext{74.2} & \ext{--} & \ext{68.5} & \ext{--} & \ext{--} & \ext{--} & \ext{--} \\
\ext{EGSDE$^\dagger$} & \ext{1200} & \ext{--} & \ext{--} & \ext{--} & \ext{--} & \ext{51.0} & \ext{--} & \ext{50.4} & \ext{--} & \ext{--} & \ext{--} & \ext{--} \\
\ext{DDIB (Teacher)} & \ext{160} & \ext{--} & \ext{--} & \ext{--} & \ext{--} & \ext{38.9} & \ext{30.3} & \ext{38.6} & \ext{13.2} & \ext{--} & \ext{--} & \ext{--} \\
\ext{IBCD$^\dagger$} & \ext{1} & \ext{--} & \ext{--} & \ext{--} & \ext{--} & \ext{44.8} & \ext{28.4} & \ext{46.1} & \ext{16.7} & \ext{--} & \ext{--} & \ext{--} \\
\midrule
\method & 1 & 81.5 & \second{76.6} & \second{33.1} & \second{116.8} & \second{37.6} & \second{20.9} & \second{37.7} & 20.8 & 74.5 & \second{51.0} & \second{55.1} \\
\method{} ($K=5$) & 5 & \second{80.9} & \best{75.9} & \best{32.6} & \best{114.5} & \best{36.8} & \best{20.4} & \best{36.9} & \second{20.2} & \second{73.6} & \best{50.2} & \best{54.2} \\
\bottomrule
\end{tabular*}
\endgroup
\caption{Direction-wise FID; lower is better. Gray rows report partial published results under their original evaluation protocols and are provided for reference only; source and configuration details are given in the surrounding text. The $\dagger$ symbol denotes the realism-prioritized configuration used in the cited comparison. All other baseline results are obtained from our reimplementations using identical data splits and metric code. Among the non-gray rows, the best and second-best result in each column are shown in boldface and with an underline, respectively.}
\label{tab:supp-main-fid}
\end{table*}


\clearpage
\onecolumn
\section{Additional Qualitative Results}
\label{sec:supp-qualitative}

Figures~\ref{fig:supp-qualitative-dc}--\ref{fig:supp-qualitative-masw} provide additional comparisons using identical source images across methods. Each figure is shown separately at a large scale to preserve local visual details.

\begin{center}
\includegraphics[width=0.96\textwidth,height=0.78\textheight,keepaspectratio]{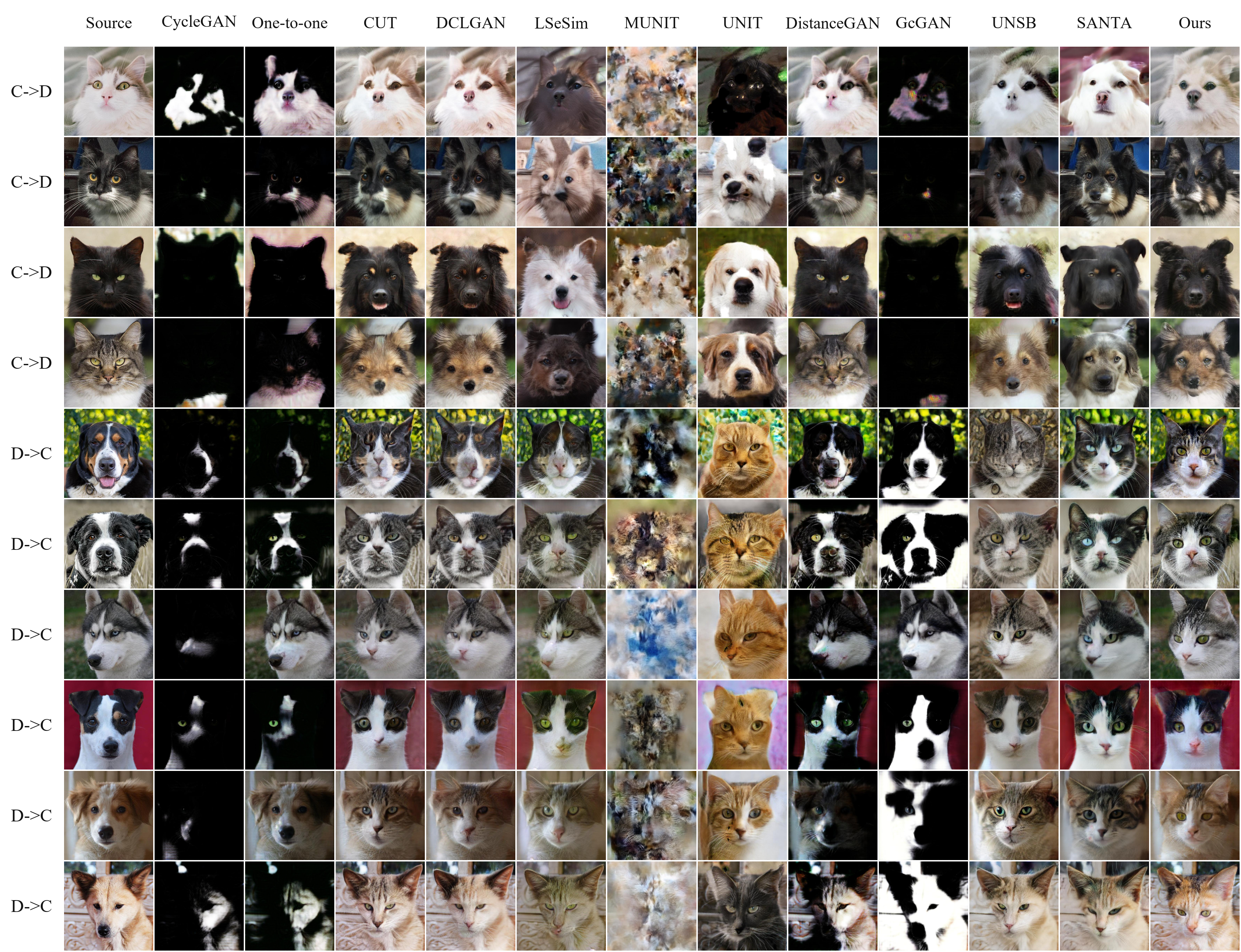}
\captionof{figure}{Additional qualitative comparisons on Cat$\leftrightarrow$Dog using identical source images across methods.}
\label{fig:supp-qualitative-dc}
\end{center}

\clearpage

\begin{center}
\includegraphics[width=0.96\textwidth,height=0.82\textheight,keepaspectratio]{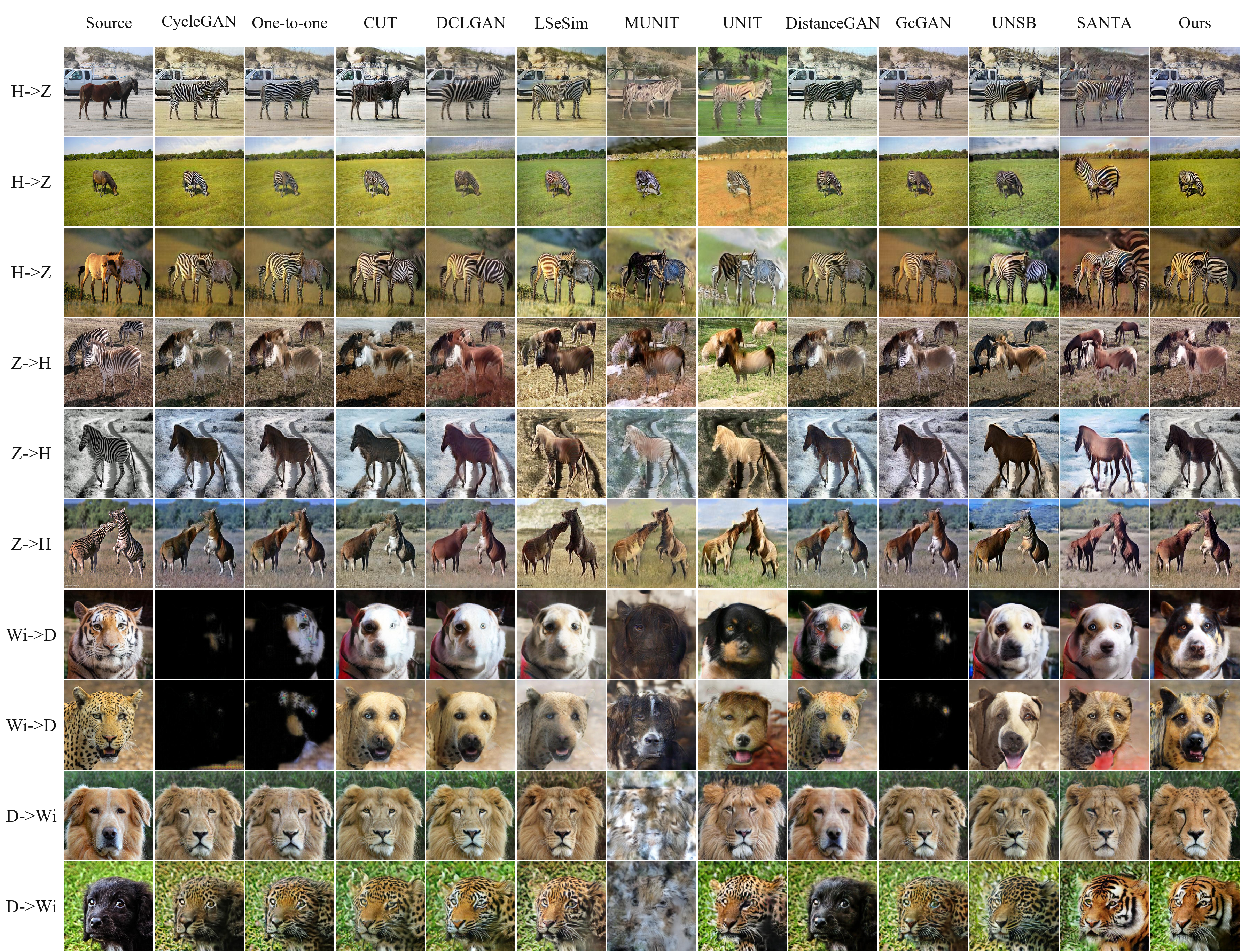}
\captionof{figure}{Additional qualitative comparisons on Horse$\leftrightarrow$Zebra and Wild$\leftrightarrow$Dog using identical source images across methods.}
\label{fig:supp-qualitative-hzwid}
\end{center}

\clearpage

\begin{center}
\includegraphics[width=0.96\textwidth,height=0.82\textheight,keepaspectratio]{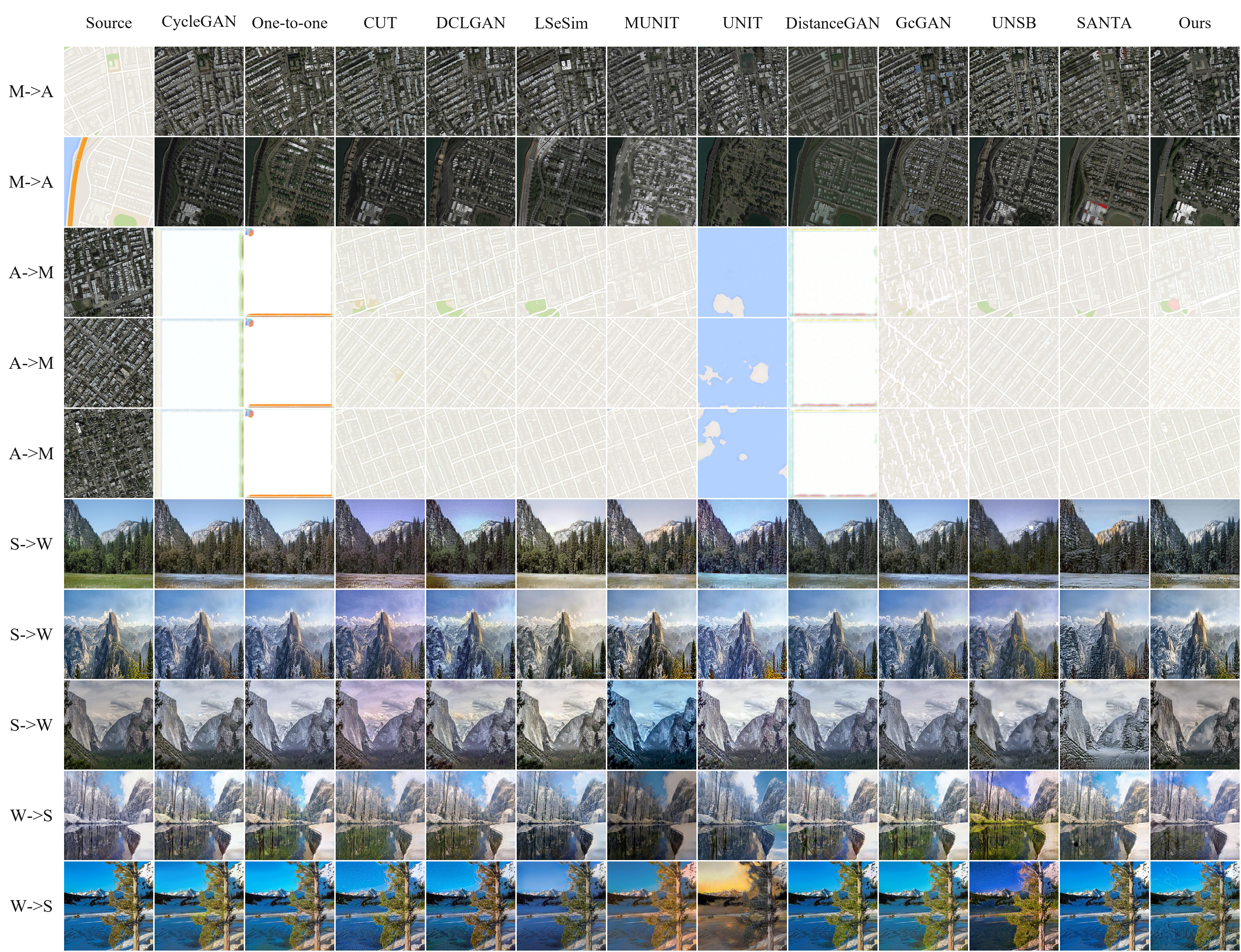}
\captionof{figure}{Additional qualitative comparisons on Aerial$\leftrightarrow$Map and Summer$\leftrightarrow$Winter using identical source images across methods.}
\label{fig:supp-qualitative-masw}
\end{center}

\clearpage
\twocolumn

\section{Complete Cross-Dataset Ablations}
\label{sec:supp-ablations}

The main paper shows Horse$\leftrightarrow$Zebra together with ten-direction averages for compactness. Here we report every direction. Unless otherwise specified, the training-objective, field-organization, and representation ablations are evaluated with NFE=5, matching the corresponding main-paper ablations.

\subsection{Training Objectives}
Table~\ref{tab:supp-component} isolates the effect of adversarial endpoint learning, selfFM, cycle closure, and RPV. Independent-pair rectified-flow training is substantially weaker. Learning source-conditioned endpoints yields the largest improvement, selfFM and cycle closure provide complementary gains, and RPV gives the best overall averages.

\begin{table*}[!t]
\centering
\begingroup
\fontsize{7.2}{8.4}\selectfont
\setlength{\tabcolsep}{0pt}
\renewcommand{\arraystretch}{1.05}

\textbf{(a) FID}\par\vspace{2pt}
\begin{tabular*}{\textwidth}{@{\extracolsep{\fill}}l *{11}{c}@{}}
\toprule
Setting & S$\to$W & W$\to$S & H$\to$Z & Z$\to$H & C$\to$D & D$\to$C & Wi$\to$D & D$\to$Wi & A$\to$M & M$\to$A & Avg. \\
\midrule
Independent-pair RF & 133.0 & 125.6 & 111.8 & 186.3 & 99.2 & 63.2 & 103.2 & 60.3 & 146.1 & 106.6 & 113.5 \\
Endpoint Adv. & 93.5 & 86.7 & 55.2 & 143.1 & 50.9 & 31.3 & 53.7 & 28.7 & 92.3 & 63.3 & 69.9 \\
+ selfFM & 87.8 & 81.5 & 43.7 & 133.4 & 43.7 & 25.5 & 45.1 & 23.0 & 82.1 & 54.5 & 62.0 \\
+ Cycle & 87.5 & 82.2 & 45.6 & 129.0 & 43.3 & 25.8 & 44.7 & 22.3 & 82.8 & 54.0 & 61.7 \\
+ selfFM + Cycle & \second{83.7} & \second{78.1} & \best{32.4} & \second{118.8} & \second{38.5} & \second{21.9} & \second{39.0} & \best{20.1} & \second{76.4} & \best{50.0} & \second{55.9} \\
+ RPV (full) & \best{80.9} & \best{75.9} & \second{32.6} & \best{114.5} & \best{36.8} & \best{20.4} & \best{36.9} & \second{20.2} & \best{73.6} & \second{50.2} & \best{54.2} \\
\bottomrule
\end{tabular*}

\vspace{5pt}
\textbf{(b) KID$\times100$}\par\vspace{2pt}
\begin{tabular*}{\textwidth}{@{\extracolsep{\fill}}l *{11}{c}@{}}
\toprule
Setting & S$\to$W & W$\to$S & H$\to$Z & Z$\to$H & C$\to$D & D$\to$C & Wi$\to$D & D$\to$Wi & A$\to$M & M$\to$A & Avg. \\
\midrule
Independent-pair RF & 5.549 & 5.123 & 5.440 & 8.715 & 4.399 & 3.058 & 4.722 & 2.683 & 7.313 & 4.589 & 5.159 \\
Endpoint Adv. & 3.499 & 3.173 & 2.890 & 6.565 & 2.149 & 1.408 & 2.272 & 1.203 & 4.213 & 2.439 & 2.981 \\
+ selfFM & 3.119 & 2.863 & 1.810 & 5.665 & 1.699 & 1.098 & 1.782 & 0.923 & 3.633 & 1.979 & 2.457 \\
+ Cycle & 3.099 & 2.893 & 1.890 & 5.485 & 1.669 & 1.118 & 1.752 & 0.893 & 3.683 & 1.949 & 2.443 \\
+ selfFM + Cycle & \second{2.879} & \second{2.663} & \best{1.330} & \second{5.045} & \second{1.409} & \second{0.918} & \second{1.442} & \best{0.773} & \second{3.273} & \best{1.729} & \second{2.146} \\
+ RPV (full) & \best{2.749} & \best{2.573} & \second{1.340} & \best{4.765} & \best{1.299} & \best{0.858} & \best{1.322} & \second{0.783} & \best{3.113} & \second{1.739} & \best{2.054} \\
\bottomrule
\end{tabular*}
\endgroup
\caption{Complete component ablation across all ten directions. Lower is better for both metrics.}
\label{tab:supp-component}
\end{table*}

\subsection{Velocity-Field Organization}
Table~\ref{tab:supp-field-organization} separates parameter sharing from temporal organization. Two direction-specific fields are weakest. A shared backbone with two output heads recovers much of the performance, while a discrete direction token remains inferior to continuous time conditioning. The shared time-conditioned field is best in every reported direction and on both averages under the NFE=5 ablation setting.

\begin{table*}[!t]
\centering
\begingroup
\fontsize{7.2}{8.4}\selectfont
\setlength{\tabcolsep}{0pt}
\renewcommand{\arraystretch}{1.05}

\textbf{(a) FID}\par\vspace{2pt}
\begin{tabular*}{\textwidth}{@{\extracolsep{\fill}}l *{11}{c}@{}}
\toprule
Velocity-field organization & S$\to$W & W$\to$S & H$\to$Z & Z$\to$H & C$\to$D & D$\to$C & Wi$\to$D & D$\to$Wi & A$\to$M & M$\to$A & Avg. \\
\midrule
Two direction-specific fields & 115.4 & 108.7 & 42.2 & 142.9 & 58.6 & 36.7 & 61.9 & 31.8 & 96.4 & 67.5 & 76.2 \\
Shared backbone, two output heads & 89.6 & 84.1 & 32.8 & 116.1 & 45.2 & 25.7 & 46.4 & 24.9 & 83.0 & 56.8 & 60.5 \\
Shared field with direction token & 91.8 & 86.5 & 33.7 & 118.2 & 46.8 & 27.1 & 48.3 & 25.8 & 85.7 & 58.4 & 62.2 \\
Shared time-conditioned field (ours) & \best{80.9} & \best{75.9} & \best{32.6} & \best{114.5} & \best{36.8} & \best{20.4} & \best{36.9} & \best{20.2} & \best{73.6} & \best{50.2} & \best{54.2} \\
\bottomrule
\end{tabular*}

\vspace{5pt}
\textbf{(b) KID$\times100$}\par\vspace{2pt}
\begin{tabular*}{\textwidth}{@{\extracolsep{\fill}}l *{11}{c}@{}}
\toprule
Velocity-field organization & S$\to$W & W$\to$S & H$\to$Z & Z$\to$H & C$\to$D & D$\to$C & Wi$\to$D & D$\to$Wi & A$\to$M & M$\to$A & Avg. \\
\midrule
Two direction-specific fields & 4.180 & 3.920 & 1.720 & 5.860 & 2.310 & 1.450 & 2.480 & 1.200 & 4.810 & 2.720 & 3.065 \\
Shared backbone, two output heads & 3.150 & 2.980 & 1.380 & 4.910 & 1.650 & 1.020 & 1.710 & 0.910 & 3.620 & 2.030 & 2.336 \\
Shared field with direction token & 3.280 & 3.090 & 1.430 & 5.020 & 1.740 & 1.090 & 1.820 & 0.960 & 3.790 & 2.110 & 2.433 \\
Shared time-conditioned field (ours) & \best{2.749} & \best{2.573} & \best{1.340} & \best{4.765} & \best{1.299} & \best{0.858} & \best{1.322} & \best{0.783} & \best{3.113} & \best{1.739} & \best{2.054} \\
\bottomrule
\end{tabular*}
\endgroup
\caption{Complete velocity-field organization ablation across all ten directions. Lower is better for both metrics. All variants use the same training objectives and are evaluated with NFE=5.}
\label{tab:supp-field-organization}
\end{table*}

\subsection{Inference Budget}
Table~\ref{tab:supp-nfe} evaluates the same trained field with different Euler step counts. NFE=1 already achieves results close to the best measured averages. NFE=2 provides most of the multi-step gain, while NFE=5 yields a smaller additional improvement. The close one-, two-, and five-step results are consistent with stable boundary and intermediate predictions over the integration range used in practice.

\begin{table*}[!t]
\centering
\begingroup
\fontsize{7.2}{8.4}\selectfont
\setlength{\tabcolsep}{0pt}
\renewcommand{\arraystretch}{1.05}
\textbf{(a) FID}\par\vspace{2pt}
\begin{tabular*}{\textwidth}{@{\extracolsep{\fill}}l *{11}{c}@{}}
\toprule
NFE & S$\to$W & W$\to$S & H$\to$Z & Z$\to$H & C$\to$D & D$\to$C & Wi$\to$D & D$\to$Wi & A$\to$M & M$\to$A & Avg. \\
\midrule
1 & 81.5 & 76.6 & 33.1 & 116.8 & 37.6 & 20.9 & 37.7 & 20.8 & 74.5 & 51.0 & 55.1 \\
2 & \best{80.6} & \second{76.4} & \best{32.4} & \second{115.3} & \best{36.5} & \second{20.6} & \best{36.7} & \best{20.0} & \second{73.9} & \best{50.1} & \second{54.3} \\
5 & \second{80.9} & \best{75.9} & \second{32.6} & \best{114.5} & \second{36.8} & \best{20.4} & \second{36.9} & \second{20.2} & \best{73.6} & \second{50.2} & \best{54.2} \\
10 & 82.9 & 78.4 & 35.8 & 121.7 & 39.8 & 22.5 & 40.2 & 22.1 & 77.1 & 52.9 & 57.3 \\
\bottomrule
\end{tabular*}

\vspace{5pt}
\textbf{(b) KID$\times100$}\par\vspace{2pt}
\begin{tabular*}{\textwidth}{@{\extracolsep{\fill}}l *{11}{c}@{}}
\toprule
NFE & S$\to$W & W$\to$S & H$\to$Z & Z$\to$H & C$\to$D & D$\to$C & Wi$\to$D & D$\to$Wi & A$\to$M & M$\to$A & Avg. \\
\midrule
1 & 2.820 & \second{2.590} & 1.390 & 4.860 & 1.345 & 0.895 & 1.410 & 0.820 & 3.180 & 1.760 & 2.107 \\
2 & \best{2.734} & 2.593 & \best{1.328} & \second{4.810} & \best{1.289} & \second{0.870} & \best{1.314} & \best{0.773} & \second{3.135} & \second{1.745} & \second{2.059} \\
5 & \second{2.749} & \best{2.573} & \second{1.340} & \best{4.765} & \second{1.299} & \best{0.858} & \second{1.322} & \second{0.783} & \best{3.113} & \best{1.739} & \best{2.054} \\
10 & 2.869 & 2.733 & 1.530 & 5.185 & 1.449 & 0.953 & 1.492 & 0.863 & 3.323 & 1.879 & 2.228 \\
\bottomrule
\end{tabular*}
\endgroup
\caption{Complete Euler-step ablation across all ten directions. Lower is better for both metrics.}
\label{tab:supp-nfe}
\end{table*}

\subsection{RPV Design}
Table~\ref{tab:supp-encoder} compares frozen encoder representations \cite{simonyan2015vgg,caron2021dino,oquab2023dinov2}. No encoder dominates every individual direction, but VGG16 \texttt{relu5\_3} gives the best ten-direction average, while DINO-based representations remain competitive. Table~\ref{tab:supp-param} studies the finite-difference interval and RPV weight. Nearby parameter choices produce similar averages, supporting the selected $\Delta=0.1$ and $\lambda_{\mathrm{RPV}}=0.05$.

\begin{table*}[!t]
\centering
\begingroup
\fontsize{7.2}{8.4}\selectfont
\setlength{\tabcolsep}{0pt}
\renewcommand{\arraystretch}{1.05}
\textbf{(a) FID}\par\vspace{2pt}
\begin{tabular*}{\textwidth}{@{\extracolsep{\fill}}l *{11}{c}@{}}
\toprule
Encoder feature & S$\to$W & W$\to$S & H$\to$Z & Z$\to$H & C$\to$D & D$\to$C & Wi$\to$D & D$\to$Wi & A$\to$M & M$\to$A & Avg. \\
\midrule
No RPV & 83.7 & 78.1 & \second{32.4} & 118.8 & 38.5 & 21.9 & 39.0 & \second{20.1} & 76.4 & \second{50.0} & 55.9 \\
VGG16 relu2\_2 & 83.4 & 77.8 & 33.1 & 118.3 & 38.3 & 21.8 & 38.7 & 21.0 & 76.1 & 51.4 & 56.0 \\
VGG16 relu3\_3 & 82.8 & 77.3 & 33.4 & 117.2 & 37.9 & 21.4 & 38.2 & 20.8 & 75.4 & 50.9 & 55.5 \\
VGG16 relu4\_3 & 81.9 & \second{75.8} & 33.0 & 116.0 & 37.4 & 21.1 & 37.6 & 20.6 & 74.4 & \best{49.9} & 54.8 \\
VGG16 relu5\_3 & \best{80.9} & 75.9 & 32.6 & \best{114.5} & \best{36.8} & \best{20.4} & \best{36.9} & 20.2 & \second{73.6} & 50.2 & \best{54.2} \\
DINO ViT-S/16, block 6 & 82.6 & 76.9 & 33.6 & 116.9 & 37.7 & 21.2 & 38.0 & 20.5 & 75.1 & 50.7 & 55.3 \\
DINO ViT-S/16, block 12 & 82.2 & 77.1 & 33.2 & 117.3 & 37.6 & 21.3 & 37.8 & 20.7 & 74.7 & 51.0 & 55.3 \\
DINOv2 ViT-B/14, block 6 & 81.7 & 76.3 & \best{32.2} & \second{115.7} & 37.3 & \second{20.8} & \second{37.4} & \best{20.0} & 74.3 & 50.5 & 54.6 \\
DINOv2 ViT-B/14, block 12 & \second{81.5} & \best{75.6} & 32.8 & 116.1 & \second{37.2} & 20.9 & 37.5 & 20.3 & \best{73.5} & 50.6 & \second{54.6} \\
\bottomrule
\end{tabular*}

\vspace{5pt}
\textbf{(b) KID$\times100$}\par\vspace{2pt}
\begin{tabular*}{\textwidth}{@{\extracolsep{\fill}}l *{11}{c}@{}}
\toprule
Encoder feature & S$\to$W & W$\to$S & H$\to$Z & Z$\to$H & C$\to$D & D$\to$C & Wi$\to$D & D$\to$Wi & A$\to$M & M$\to$A & Avg. \\
\midrule
No RPV & 2.879 & 2.663 & \second{1.330} & 5.045 & 1.409 & 0.918 & 1.442 & \second{0.773} & 3.273 & \best{1.729} & 2.146 \\
VGG16 relu2\_2 & 2.864 & 2.653 & 1.365 & 5.010 & 1.389 & 0.913 & 1.427 & 0.818 & 3.258 & 1.809 & 2.151 \\
VGG16 relu3\_3 & 2.834 & 2.633 & 1.380 & 4.945 & 1.364 & 0.898 & 1.397 & 0.808 & 3.218 & 1.784 & 2.126 \\
VGG16 relu4\_3 & 2.794 & \second{2.568} & 1.360 & 4.855 & 1.334 & 0.888 & 1.362 & 0.798 & 3.158 & \second{1.729} & 2.085 \\
VGG16 relu5\_3 & \best{2.749} & 2.573 & 1.340 & \best{4.765} & \best{1.299} & \best{0.858} & \best{1.322} & 0.783 & \second{3.113} & 1.739 & \best{2.054} \\
DINO ViT-S/16, block 6 & 2.824 & 2.618 & 1.390 & 4.910 & 1.349 & 0.893 & 1.382 & 0.793 & 3.198 & 1.769 & 2.113 \\
DINO ViT-S/16, block 12 & 2.809 & 2.628 & 1.370 & 4.930 & 1.344 & 0.898 & 1.372 & 0.803 & 3.178 & 1.779 & 2.111 \\
DINOv2 ViT-B/14, block 6 & 2.779 & 2.591 & \best{1.328} & \second{4.835} & 1.319 & \second{0.873} & \second{1.347} & \best{0.770} & 3.143 & 1.751 & 2.074 \\
DINOv2 ViT-B/14, block 12 & \second{2.774} & \best{2.563} & 1.350 & 4.845 & \second{1.317} & 0.878 & 1.352 & 0.788 & \best{3.107} & 1.757 & \second{2.073} \\
\bottomrule
\end{tabular*}
\endgroup
\caption{Complete RPV encoder ablation across all ten directions. Lower is better for both metrics.}
\label{tab:supp-encoder}
\end{table*}

\begin{table*}[!t]
\centering
\begingroup
\fontsize{7.0}{8.2}\selectfont
\setlength{\tabcolsep}{0pt}
\renewcommand{\arraystretch}{1.05}
\textbf{(a) FID}\par\vspace{2pt}
\begin{tabular*}{\textwidth}{@{\extracolsep{\fill}}l l *{11}{c}@{}}
\toprule
Group & Value & S$\to$W & W$\to$S & H$\to$Z & Z$\to$H & C$\to$D & D$\to$C & Wi$\to$D & D$\to$Wi & A$\to$M & M$\to$A & Avg. \\
\midrule
$\Delta$ & 0.05 & 81.3 & \best{75.7} & 33.2 & \second{115.8} & \second{37.0} & 20.5 & 37.3 & \best{20.1} & 74.3 & 50.5 & 54.6 \\
 & 0.10 & \second{80.9} & \second{75.9} & \second{32.6} & \best{114.5} & \best{36.8} & \second{20.4} & \best{36.9} & \second{20.2} & \best{73.6} & \best{50.2} & \best{54.2} \\
 & 0.20 & \best{80.6} & 76.2 & \best{32.3} & 116.2 & 37.2 & \best{20.3} & \second{37.1} & 20.3 & \second{74.0} & \second{50.3} & \second{54.5} \\
 & 0.40 & 83.3 & 77.9 & 36.1 & 120.5 & 39.6 & 22.1 & 39.9 & 21.6 & 76.7 & 52.6 & 57.0 \\
\midrule
$\lambda_{\mathrm{RPV}}$ & 0.010 & 83.0 & 77.7 & 35.7 & 119.7 & 39.0 & 21.8 & 39.4 & 21.4 & 76.2 & 52.1 & 56.6 \\
 & 0.025 & 81.6 & 76.4 & 33.5 & 116.0 & 37.4 & 20.8 & 37.7 & 20.4 & 74.5 & 50.7 & 54.9 \\
 & 0.050 & \best{80.9} & \second{75.9} & \best{32.6} & \second{114.5} & \best{36.8} & \best{20.4} & \best{36.9} & \second{20.2} & \best{73.6} & \best{50.2} & \best{54.2} \\
 & 0.100 & \second{81.3} & \best{75.7} & \second{33.4} & \best{114.2} & \second{37.1} & \second{20.6} & \second{37.3} & \best{20.1} & \second{74.2} & \second{50.3} & \second{54.4} \\
 & 0.200 & 84.4 & 78.9 & 38.2 & 121.6 & 40.6 & 22.8 & 40.9 & 22.0 & 77.8 & 53.4 & 58.1 \\
\bottomrule
\end{tabular*}

\vspace{5pt}
\textbf{(b) KID$\times100$}\par\vspace{2pt}
\begin{tabular*}{\textwidth}{@{\extracolsep{\fill}}l l *{11}{c}@{}}
\toprule
Group & Value & S$\to$W & W$\to$S & H$\to$Z & Z$\to$H & C$\to$D & D$\to$C & Wi$\to$D & D$\to$Wi & A$\to$M & M$\to$A & Avg. \\
\midrule
$\Delta$ & 0.05 & 2.767 & \best{2.565} & 1.370 & \second{4.840} & \second{1.311} & 0.863 & 1.342 & \best{0.777} & 3.148 & 1.753 & 2.074 \\
 & 0.10 & \second{2.749} & \second{2.573} & \second{1.340} & \best{4.765} & \best{1.299} & \second{0.858} & \best{1.322} & \second{0.783} & \best{3.113} & \best{1.739} & \best{2.054} \\
 & 0.20 & \best{2.739} & 2.585 & \best{1.325} & 4.855 & 1.317 & \best{0.854} & \second{1.332} & 0.789 & \second{3.133} & \second{1.743} & \second{2.067} \\
 & 0.40 & 2.854 & 2.663 & 1.520 & 5.125 & 1.429 & 0.938 & 1.467 & 0.843 & 3.278 & 1.859 & 2.198 \\
\midrule
$\lambda_{\mathrm{RPV}}$ & 0.010 & 2.844 & 2.653 & 1.495 & 5.065 & 1.409 & 0.923 & 1.447 & 0.828 & 3.248 & 1.829 & 2.174 \\
 & 0.025 & 2.779 & 2.595 & 1.385 & 4.850 & 1.329 & 0.876 & 1.362 & 0.793 & 3.158 & 1.764 & 2.089 \\
 & 0.050 & \best{2.749} & \second{2.573} & \best{1.340} & \second{4.765} & \best{1.299} & \best{0.858} & \best{1.322} & \second{0.783} & \best{3.113} & \best{1.739} & \best{2.054} \\
 & 0.100 & \second{2.767} & \best{2.563} & \second{1.380} & \best{4.745} & \second{1.314} & \second{0.868} & \second{1.342} & \best{0.778} & \second{3.143} & \second{1.744} & \second{2.064} \\
 & 0.200 & 2.904 & 2.708 & 1.625 & 5.195 & 1.479 & 0.963 & 1.517 & 0.858 & 3.333 & 1.899 & 2.248 \\
\bottomrule
\end{tabular*}
\endgroup
\caption{Complete RPV parameter ablation across all ten directions. Lower is better for both metrics; rankings are computed separately within the $\Delta$ and $\lambda_{\mathrm{RPV}}$ groups. When varying $\Delta$, we fix $\lambda_{\mathrm{RPV}}=0.05$;
when varying $\lambda_{\mathrm{RPV}}$, we fix $\Delta=0.1$.
Rankings are computed separately within the two groups.}
\label{tab:supp-param}
\end{table*}

\clearpage

\raggedbottom
\setlength{\bibsep}{0pt}


\def\isChecklistMainFile{}

\end{document}